\documentclass[lettersize,journal]{IEEEtran}
\usepackage{amsmath,amsfonts}
\usepackage{algorithmic}
\usepackage{algorithm}
\usepackage{array}
\usepackage[caption=false,font=footnotesize,labelfont=sf,textfont=sf]{subfig}
\usepackage{textcomp}
\usepackage{stfloats}
\usepackage{url}
\usepackage{verbatim}
\usepackage{graphicx}
\usepackage{cite}
\usepackage{multirow}

\DeclareMathAlphabet{\pazocal}{OMS}{zplm}{m}{n}
\begin{document}

\title{Spiking Neural Network Feature Discrimination Boosts Modality Fusion}
\author{Katerina Maria Oikonomou\textsuperscript{1 *}, Ioannis Kansizoglou\textsuperscript{1}, and Antonios Gasteratos\textsuperscript{1}
\thanks{\textsuperscript{1}K. M. Oikonomou, I. Kansizoglou, and A. Gasteratos are with the Department of Production and Management Engineering, Laboratory of Robotics and Automation, Democritus University of Thrace, Xanthi, Greece {\tt\footnotesize emails: \{aioikono, ikansizo, agaster\}@pme.duth.gr}
}
\thanks{This research has been co-financed by the European Health and Digital Executive Agency (HADEA), under the powers delegated by the European Commission (‘European Commission’) (project code: MASTERMINE-101091895.}

\thanks{* Correspondence: aioikono@pme.duth.gr}
}

\markboth{Journal of \LaTeX\ Class Files,~Vol.~14, No.~8, August~2021}%
{Shell \MakeLowercase{\textit{et al.}}: A Sample Article Using IEEEtran.cls for IEEE Journals}


\maketitle

\begin{abstract}
Feature discrimination is a crucial aspect in neural network design, as it directly impacts the network's ability to distinguish between classes and generalize across diverse datasets.
The accomplishment of achieving high-quality feature representations ensures high intra-class separability and poses one of the most challenging research directions.
While convetional deep neural networks (DNNs) rely on complex transformations and very deep networks to come up with meaningful feature representations, they usually require days of training and consume significant energy amounts.
To this end, spiking neural networks (SNNs) offer a promising alternative.
SNN's ability to capture temporal and spatial dependencies renders them particularly suitable for complex tasks, where multi-modal data are required.
In this paper, we propose a feature discrimination approach for multi-modal learning with SNNs, focusing on audio-visual data.
We employ deep spiking residual learning for visual modality processing and a simpler yet efficient spiking network for auditory modality processing.
Lastly, we deploy a spiking multilayer perceptron for modality fusion.
We present our findings and evaluate our approach against similar works in the field in classification challenges.
To the best of our knowledge, this is the first work investigating feature discrimination in SNNs.

\end{abstract}


\section{Introduction}
\IEEEPARstart{A}{rtificial} intelligent (AI) models demonstrate a tremendous demand for diverse data to produce more accurate and robust systems. 
Yet, training such models with large datasets increases computational complexity and power consumption that on average can reach up to 250 watts (W)~\cite{7723730}.
Meanwhile, contemporary approaches studied a correlation between a neural network's performance and its depth, measured in terms of hidden layers, where deeper neural networks tend to process more complex patterns with higher representation capacities~\cite{sze2017efficient,szegedy2015going,touvron2021going}.
However, these representations come with a large number of parameters and high power consumption.
Furthermore, studies have revealed the degradation problem, which indicates that deeper networks can not guarantee better performance~\cite{he2016deep,he2015convolutional,srivastava2015highway}.
To alleviate that, residual networks effectively address this issue at very deep networks~\cite{he2016deep,shafiq2022deep,zhang2017residual}.
However, deep networks' energy efficiency is still far from optimal despite their benefits in learning processes.
Meanwhile, inspired by the brain's efficient information processing capabilities, the design of spiking neural networks (SNNs) has emerged as a promising solution to address the limitations of deep networks.
SNNs have gained great interest over the past years due to their high biological possibility, temporal dynamics, and low power consumption~\cite{roy2019towards}~\cite{hussaini2024spiking}.
SNNs were first introduced in the late 1990s~\cite {gerstner1999spike} and ever since have been successfully applied in numerous tasks, including control~\cite{oik2023hybrid,oikonomou2023hybrid,9340948}, detection~\cite{fan2024sfod} and classification~\cite{5585775,ponulak2010supervised,9674199}.

\begin{figure}
    \centering
    \includegraphics[scale=0.5]{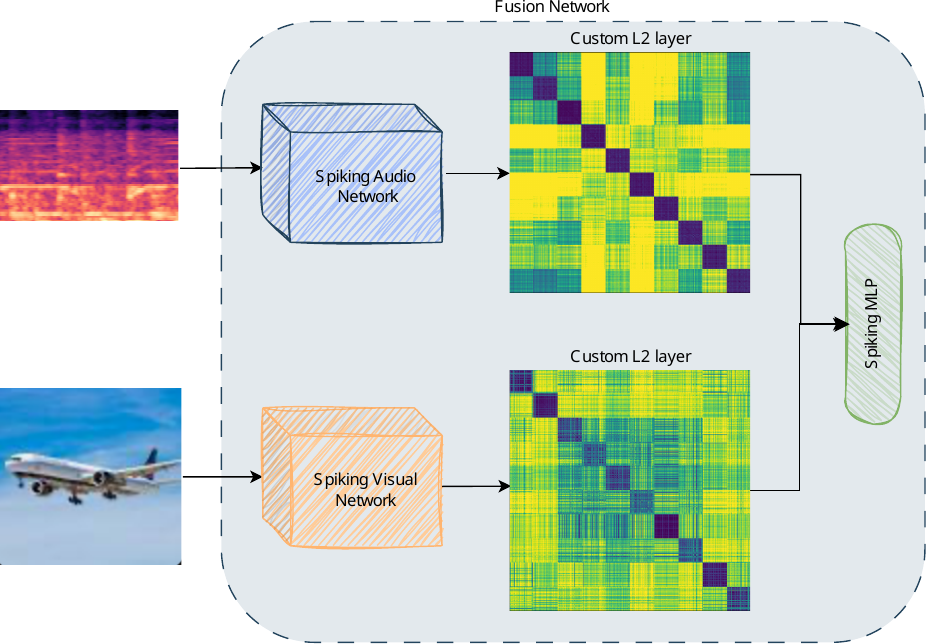}
    \caption{Overview of the proposed multi-modal spiking fusion architecture for audio-visual classification, integrating feature discrimination with spiking neural networks.}
    \label{fig:overview}
\end{figure}

Recent approaches incorporate prior knowledge from DNNs in SNNs research.
This includes techniques for transforming artificial neural networks (ANNs) into SNNs~\cite{ding2021optimal,10285018}, applying deep learning tools, such as attention mechanisms~\cite{yao2023attention,liu2022event} multi-layer perceptrons (MLPs)~\cite{oikonomou2023bio}, and the conversion of well-established deep architectures, \textit{e.g.}, SNN implementation of VGG and ResNet~\cite{lee2020enabling,sengupta2019going,shi2024spikingresformer}.
An important property direction of DNNs constitutes their feature discrimination capacities, directly estimating the network's ability to discriminate classes, \textit{i.e.}, extracting features with high intra-class compactness and inter-class separability.
The above fact, is proven highly beneficial in cascaded tasks and multi-modal models, considering that, multi-modal data has drawn a lot of interest in recent classification problems.
Although several neural networks have been successfully trained with uni-modal data, the enhanced representations of multi-modal data lead to improved performance in many complex tasks.
This aligns with findings from neuroscience, where cross-modal interactions are fundamental to perception~\cite{shimojo2001sensory}.
Visual signals are modulated by other modalities, \textit{i.e.}, auditory, tactile, or vestibular, thus reflecting the natural integration observed in biological systems.
Having said that, it becomes evident that a combination of feature discrimination approaches with multi-modal datasets in multi-modal SNNs~\cite{9746865,8482490,10293172} can emerge as a promising solution, leveraging the energy efficiency of SNNs with the representational power of multi-modal learning.
In summary, the contributions of the proposed method are as follows:
\begin{itemize}
    \item the introduction of an L2 normalization layer after the last hidden one to enhance feature discrimination in uni-modal SNN architectures;
    \item the design of a novel spiking ResNet approach for the visual uni-modal architecture, \textit{viz.}, L2-ActSpikeNet, and an efficient spiking model for processing the audio modality;
    \item the design of a spiking MLP (SMLP) architecture for modalities fusion and efficient classification as shown in Fig.~\ref{fig:overview}.
\end{itemize}
To the best of our knowledge, this is the first work attempting to introduce a feature discrimination approach in SNNs, which is further evaluated in the practical task of audio-visual classification.

\section{Related Works}
In this section, we explore recent research trends in spiking architectures, multi-modal classification techniques, and feature discrimination approaches.
Building on the aforementioned research fields, the design of our multi-modal spiking architecture has been shaped, combining the strengths of feature discrimination techniques with the efficiency of multi-modal integration.

\subsection{Spiking Neural Network}

SNNs have been extensively studied thanks to their high biological plausibility and the remarkably low computational complexities achieved with neuromorphic processors~\cite{davies2021advancing,massa2020efficient}. 
They have been designed with bio-plausible learning rules~\cite{iakymchuk2015simplified,hao2020biologically,kheradpisheh2018stdp}, by adapting prior knowledge of convolutional neural networks (CNNs) or through several ANN2SNN conversion approaches~\cite{bu2023optimal,wang2024universal,wang2023new}.
SNNs have also shown remarkable performance with backpropagation approaches~\cite{deng2022temporal,guo2023membrane, lee2020enabling} as a result of extensive research on efficient learning approaches~\cite{meng2023towards, wei2023temporal}.
Spiking networks such as spiking VGG~\cite{lee2020enabling,sengupta2019going} and spiking ResNet~\cite{shi2024spikingresformer,zheng2021going,fang2021deep,zhou2023spikingformer,hu2021spiking}, achieve high accuracy by taking advantage of the useful characteristics of the corresponding conventional DNN architectures.
In~\cite{hu2021spiking}, an ANN2SNN conversion is performed using a residual model to scale the activations along with a compensation mechanism to minimize the discretization errors.
In~\cite{sengupta2019going}, the authors suggested an ANN2SNN conversion technique to balance the SNN's threshold for the VGG and ResNet spiking approaches. 
Other approaches propose novel backpropagation algorithms to achieve state-of-the-art results with residual networks~\cite{lee2020enabling}.
Zheng et al.~\cite{zheng2021going} replaced the batch normalization layer (BN) with a custom threshold dependent, introducing a more bio-plausing normalization approach in spiking ResNet.
Fang et al.~\cite{fang2021deep}, introduced a spike-element-wise approach in the common ResNet architecture for residual learning, ensuring identity mapping and overcoming the vanish-exploding gradients problem.

\subsection{Multi-modal classification}

Multi-modal classification, particularly in the audio-visual domain, has gained significant attention for its ability to enhance model performance by leveraging complementary cross-modal features.
In early approaches, multi-modal recognition has been realized through feature level~\cite{natarajan2012multimodal} or decision level fusion~\cite{ebrahimi2015recurrent} based on the stage where the combination of the different modalities is performed. 
Gradually, given their efficiency in classification approaches, ANNs became the most widespread selection for audio-visual classification adopting a feature-level fusion methodology~\cite{metallinou2012context,eyben2011audiovisual,8937495}.
In addition, recurrent neural networks (RNNs) were frequently used in corresponding techniques to effectively capture the temporal properties of a given task~\cite{feng2017audio,makino2019recurrent}.
Yet, despite the wide adoption of several DNN architectures in multi-modal tasks and the promising results of SNNs, the utilization of the latter in multi-modal classification tasks is still in its infancy.
To that end, an unsupervised method using multi-modal SNNs that exploits cross-modal connections and spike-timing-dependent plasticity to interpret visual and auditory data has been presented~\cite{rathi2018stdp}.
Similarly, by dynamically modifying the weights allocated to each modality, the authors in~\cite{liu2022event} introduced the first supervised multi-modal SNN intended for audio-visual classification, also employing an attention mechanism.
Meanwhile, a spiking cross-modal attention mechanism for audio-visual classification with SNN has been proposed, displaying enhanced performance~\cite{10293172}

\subsection{Feature discrimination}
Feature discrimination has gained great attention over the last few years.
Maintaining relatively high intra-class compactness compared to inter-class discrepancy has been a key focus in many learning methodologies in the field~\cite{sun2014deep, wen2016discriminative, adeli2018semi}.
Contemporary methods involve adding angular margins between classes to reinforce the loss function of DNNs to increase the separability between feature vectors. 
To achieve that, the Large-Margin Softmax Loss (L-Softmax)~\cite{liu2016large} combines a softmax function and a normalization scheme applied to the feature vectors of the final fully connected layer to produce tighter angular bounds.
Further empirical and theoretical studies demonstrated that adding hyperspherical or decoupled convolution operations to CNNs can enhance performance~\cite{liu2017deep}. 
For instance, L2-normalization combined with angular margin-based loss functions~\cite{deng2019arcface,liu2017sphereface} has significantly enhanced performance by enforcing intra-class compactness and inter-class separability on the studied hypersphere.
The abovementioned methods produce robust embeddings, while also presenting highly discriminative features.

\section{Motivation}
Even though feature discrimination holds a wide research direction in DNNs, to the best of our knowledge, there are no works with SNNs.
By examining the output space of the last hidden layer of a DNN, in a previous work of ours, we have proved that the feature vectors of the studied space are angularly discriminated~\cite{9477034} when the softmax activation function is applied to extract the classification result.
Inspired by that, in this work, we investigate how feature discrimination can be also applied in SNNs.

In particular, we form our motivation regarding feature discrimination in SNNs considering the common leaky integrate and fire (LIF) neurons described by the following equations:

\begin{equation} \label{eq:lif1}
V[t+1] = \beta V[t]+ I_{in}[t+1] 
\end{equation}
and 
\begin{equation}
V[t] > V_{th} \Rightarrow S[t+1] = 1,
\end{equation}
where $V[t]$ is the membrane potential at time-step $t$, $V_{th}$ the membrane threshold, $I_{in}$ the input current and $\beta \in [0,1]$ the membrane potential decay rate.
$S[t+1]$ is the output spike at time-step $t+1$, triggered when the membrane potential at time-step $t$ exceeds its threshold.

\textbf{Lemma 1.} \textit{Consider an SNN network with LIF neurons in the last layer and the mean squared error (MSE) spike count loss function. 
Since, SNNs inherently incorporate temporal dynamics into their computations, at time-step $t=0$, the feature vectors in the output of the last hidden layer are angularly discriminated.}

\textit{Proof:}
At time-step $t=0$, class $\mathcal{A}$ surpasses class $\mathcal{B}$ when the output neuron of class $\mathcal{A}$ generates a spike, while the corresponding output neuron of class $\mathcal{B}$ remains silent.
To that end, we want the following conditions to be satisfied, \textit{i.e.}, $S_A[0] = 1$ and $S_B[0] = 0$. 
Considering the LIF neurons' rule described in Eq.~\ref{eq:lif1}, we can write:
\begin{equation} \label{eq:Va}
V_A[1] \geq V_{th} \wedge V_B[1] < V_{th} \Rightarrow  \\
V_A[1] > V_B[1].
\end{equation}
Paying attention to Eq.~\ref{eq:lif1}, $V_A[0]=0$ and $V_B[0]=0$, while $I_{in}$ represents the output of the last hidden layer from now on referred to as logits. 
In~\cite{9477034}, we have proved that the logits of the $i$-th output neuron ($z_i$) can be described as:
\begin{equation}
z_i = \bar{a}\cdot \bar{w_i},
\end{equation}
where, $\bar{a}\in{F}$ denotes the expanded feature vector in the studied feature space $\mathcal{F}\subseteq \mathbb{R}^{n+1}$ and $\bar{w_i}\in{F}$ the $i$-th output neuron's trainable parameters.
In our case we have, $I_{in} = z_i$, thus modifying Eq.~\ref{eq:lif1} as follows:
\begin{multline}
V_A[1]=\beta V_A[0]+I_{in}[1] \Rightarrow \\
V_A[1]=0+I_{in}[1]  \Rightarrow \\
V_A[1]=\bar{a}\cdot\bar{w_A}
\end{multline}
Similarly, $V_B[1]=\bar{a}\cdot \bar{w_B}$.
From Eq.~\ref{eq:Va}, we have:
\begin{equation} \label{eq:az}
\bar{a}\cdot \bar{w_A} > \bar{a}\cdot \bar{w_B}.
\end{equation}
Following the similar approach with the Feature Space Division proof in~\cite{9477034}, it is straightforward that Eq. \ref{eq:az} ensures angular separability at time-step 0.

\textbf{Lemma 2.} \textit{Consider an SNN network with LIF neurons in the last layer and the mean squared error spike count loss function. 
Given a timestep $t$, the dominance of an output class is defined by the weighted accumulation of the dot products between the class's weights and the feature vectors among all the previous timesteps.}

\textit{Proof:}
At time-step $t = 2$, we have:
\begin{multline}\label{eq:t2}
V_A[2]=\beta V_A[1]+I_{in}[2] \Rightarrow \\
V_A[2]=\beta I_{in}[1]+I_{in}[2]  \Rightarrow \\
V_A[2]=\beta \bar{a}\cdot \bar{w_A}^{t=0}+\bar{a}\cdot \bar{w_A}^{t=1} 
\end{multline}

We can observe that, the Eq.~\ref{eq:t2} can be expanded for subsequent timesteps through a  mathematical series, as follows:
\begin{equation} \label{eq:ser}
V_A[t] = \sum^{t-1}_{k=0} \beta^k(\bar{a}[k]\cdot\bar{w_A}[k]).
\end{equation}

Thus, class $\mathcal{A}$ surpasses class $\mathcal{B}$ if $V_A[t] > V_B[t]$, which applies if and only if:
\begin{equation}\label{eq:vineq}
\sum^{t-1}_{k=0} \beta^k(\bar{a}[k]\cdot\bar{w_A}[k]) > \sum^{t-1}_{k=0} \beta^k(\bar{a}[k]\cdot\bar{w_B}[k]).
\end{equation}

Diving into Eq. \ref{eq:vineq}, we focus our attention on the optimal case in which an SNN is capable of confidently classifying each sample into the correct class.
The above fact implies that the SNN can correctly classify each sample regardless of the timestep, \textit{i.e.}, $\forall t$.
Considering that $\beta$ is common for all classes and $\beta > 0$, we measure the accumulated cosine similarities between a given pair of feature vectors of two distinctive samples among all the timesteps.
This examination lies on the motivation that $\forall t$ the higher dot product values of class $\mathcal{A}$ in Eq. \ref{eq:vineq} leads to higher cosine similarities of the feature vectors with the weight $\bar{w}_\mathcal{A}$.
Therefore, feature vectors of the same class are angularly concentrated $\forall t$, similarly to the case of $t=1$ from Lemma 1.
The above motivation is exploited in our methodology and empirically studied in our experiments.

\section{Methods}
In this section, we propose a novel fusion approach for audio-visual classification, exploiting feature discrimination's critical role in improving modalities fusion performance. 
The Spiking MLP (SMLP) model combines SNNs with MLP to achieve high classification accuracy on multi-modal datasets.
In the following subsections, we outline the architecture of our proposed multi-modal classification framework.
Section~\ref{sec:snnneuron} presents the network's spiking neurons, Section~\ref{sec:norm} presents the implementation of the L2 normalization used in the spiking networks, Section~\ref{sec:visualft} presents the visual features extraction with our L2-ActSpikeNet architecture, Section~\ref{sec:audioft} is the designed spiking network for the audio modality, and Section~\ref{sec:fusion} is the SMLP network for the fusion of the uni-modal extractors.

\subsection{Spiking Neuron Model}\label{sec:snnneuron}
Inspired by the neuronal dynamics met in biological brains, SNNs employ spiking neurons as their fundamental computational units. 
Spiking neurons communicate through discrete events, known as spikes, which occur when their membrane potential surpasses a specific threshold.
Thanks to their event-driven behaviour, spiking neurons are especially well-suited for tasks involving sequential or event-based data, as they can effectively encode temporal and spatial information.
In this work, we used the first-order LIF neuron model.
The neuronal dynamics that describe the LIF neuron is expressed in Eq.~\ref{eq:lif1}.
The non-differentiability of the spiking neuron makes their training with gradient-based optimization techniques extremely challenging. 
To address that, approaches approximate the spike with custom differentiable functions.
In our approach, we use different gradient approximations based on the data.
Further analysis of the learning approach is described in Section~\ref{subsec:cifar-dat}.

\begin{figure*}
    \centering
    \includegraphics[scale=0.5]{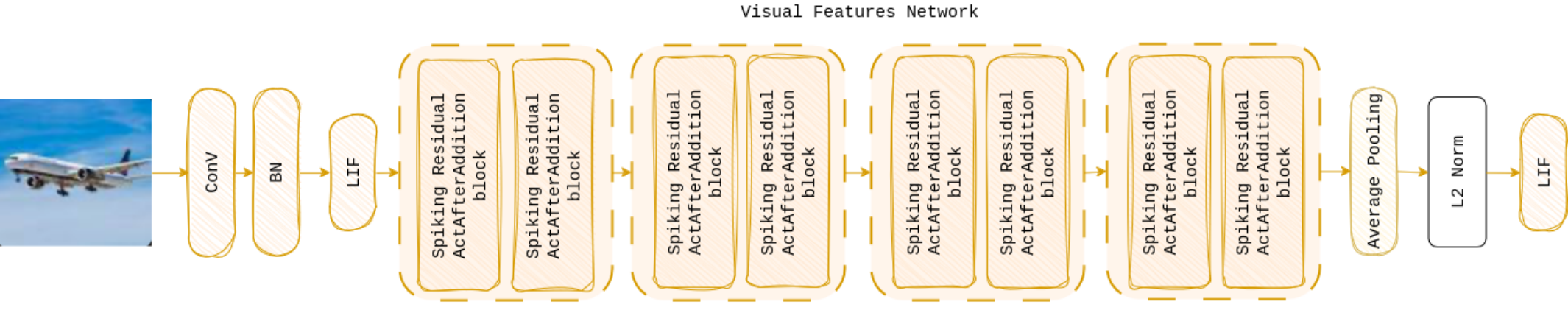}
    \caption{Architecture of the L2-ActSpikeNet visual feature extraction network, incorporating residual learning with spiking neurons and our proposed L2 normalization layer for enhanced feature discrimination.}
    \label{fig:visualnet}
\end{figure*}

\subsection{L2 Normalization}\label{sec:norm}
As already discussed in previous sections, the spiking nature of both visual and audio models introduces temporal dynamics, and thus the embeddings represent aggregated features across timesteps.
After processing an input $S^l[T]$ through the spiking layers over $T$ timesteps, the aggregated features among timesteps are transformed through the L2 normalization.
For ${z_t}\in\mathbb{R}^{d}$, where $z$ the feature vector at timestep $t$ after average pooling and $\textbf{Z}=[{z_1},{z_2},..,{z_T}]\in\mathbb{R}^{T\times d}$ the matrix of the feature vectors of a given input among all timesteps, the averaged feature vector after temporal processing $z_{av}\in\mathbb{R}^{d}$ is computed as:
\begin{equation}
{z}_{av} = \frac{1}{T} \sum^{T}_{t=1}{z_t}.
\end{equation}
This vector is then normalized using the L2 norm as follows:
\begin{equation}
\textbf{z} = \frac{{z_{av}}}{\lVert {z}_{av} \rVert_2} \quad, 
\end{equation}
\begin{equation*}
\textrm{where} \quad 
\lVert {z}_{av} \rVert_2 = \sqrt{ \sum^{d}_{i=1}({z}_{av},i})^2.
\end{equation*}
Meanwhile, each class weight vector ${w}_j\in\mathbb{R}^{d}$ of the learnable weight matrix $\textbf{W}\in\mathbb{R}^{d\times C}$, where $C$ the number of classes, is also normalized as follows:
\begin{equation}
\hat{{w}}_j = \frac{{w_{j}}}{\lVert {w}_{j} \rVert_2} \quad, 
\end{equation}
\begin{equation*}
\textrm{where} \quad \lVert {w}_{j} \rVert_2 = \sqrt{ \sum^{d}_{i=1}({w}_{j},i})^2.
\end{equation*}

Accordingly, the logits $\textbf{L}\in\mathbb{R}^{B\times C}$, where $B$ the batch, are computed as follows:
\begin{equation}
\textbf{L}_{ij}=\hat{\textbf{z}}_i^T \hat{\textbf{w}}_j.
\end{equation}
where $\hat{\textbf{z}}_i^T$ the normalized embedding for the $i$-th sample in the batch and $ \hat{\textbf{w}}_j$ the normalized weight vector for the $j$-th class.

\begin{figure}
    \centering
    \includegraphics[scale=0.5]{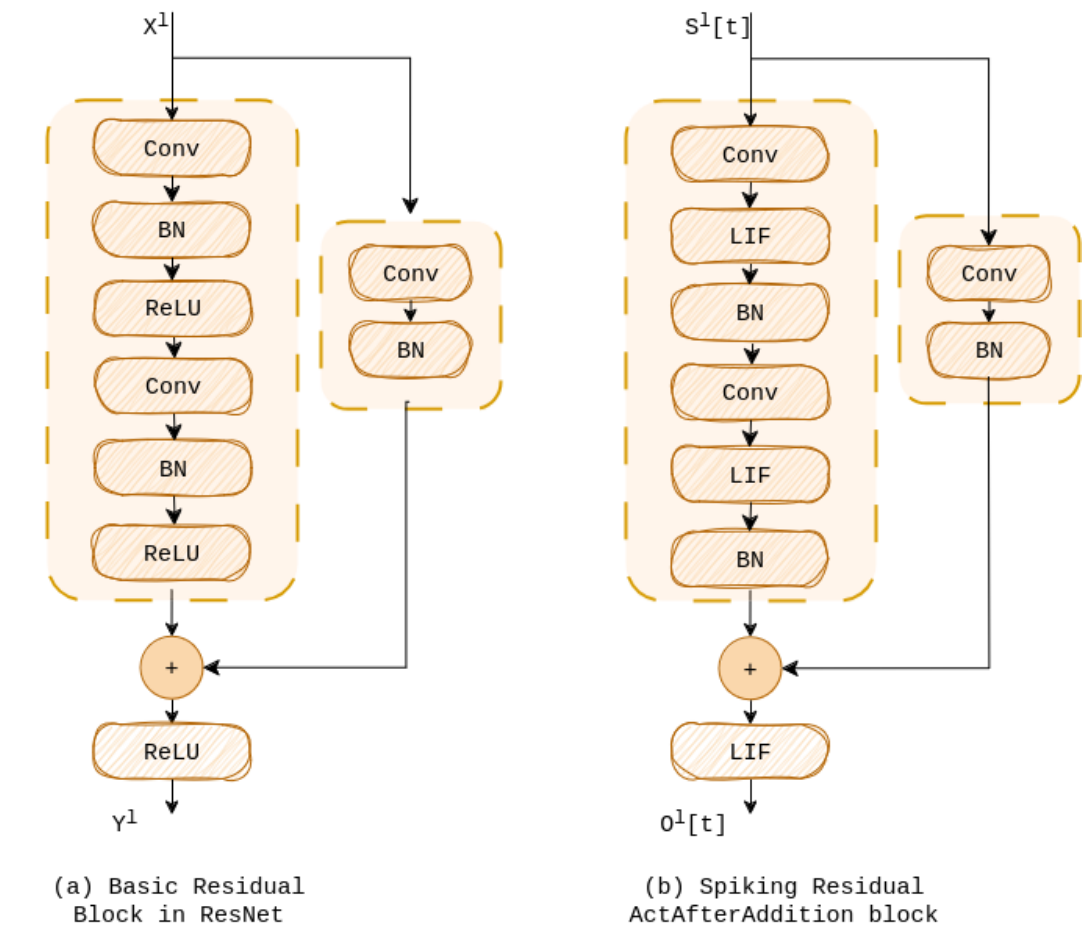}
    \caption{Comparison of Fig. A the standard residual block in conventional ResNet and Fig. B the proposed ActAfterAddition block in our L2-ActSpikeNet approach.}
    \label{fig:resnet}
\end{figure}

\subsection{Visual Features with L2-ActSpikeNet}\label{sec:visualft}
The extraction of the visual features is achieved by exploiting the advantages of SNNs.
More specifically, we designed the L2-ActSpikeNet architecture, which utilizes residual connections and the proposed L2 normalization ensuring robust feature learning, while maintaining the computational capabilities of a spiking network.
This architecture processes input images across multiple timesteps, preserving spatio-temporal information. 
Fig.~\ref{fig:resnet} illustrates the basic block of both the conventional ResNet and the proposed Spiking ResNet basic block, \textit{viz.}, the ActAfterAddition block. 
Here, $X^l$ denotes the input vector of the $l$-th residual block and $Y^l$ the corresponding output, while $S^l[t]$ is the spiking input of the $l$-th residual block and $O^l[t]$ the spiking output of the corresponding spiking block.
The convolutional layer of the conventional residual block is followed by a batch normalization (BN) layer and a rectified linear unit (ReLU) activation function, while in the spiking residual block, the convolutional layer is followed by a LIF layer and a BN one.
Many spiking ResNet approaches with ANN2SNN conversions replace the ReLU activation function with the spiking neuron layer.
Subsequently, in very deep networks, like ResNet, the spiking neuron is handled as an activation function.
While ReLU introduces non-linearity to the ANNs, LIF neurons are designed to integrate temporal information over multiple timesteps and generate spikes when their threshold membrane potential is exceeded. 
Our approach is based on spiking ResNet18.
Furthermore, we place the LIF neurons immediately after the convolutional layer ensuring that the temporal dynamics of the neurons are maintained within the deep network.
While LIF neurons are designed to process raw input, the BN layer can interfere with the cumulative behaviour of the membrane potential.
Additionally, in biological neurons, the ``normalization'' or regulation of the spiking activity typically occurs at later stages in the biological processing hierarchy, \textit{e.g.}, synaptic scaling, homeostasis.
The above fact is mimicked by the proposed approach by applying LIF neurons directly after the convolutional layer and our custom L2 normalization at the last layers.
\begin{figure*}
    \centering
    \includegraphics[scale=0.6]{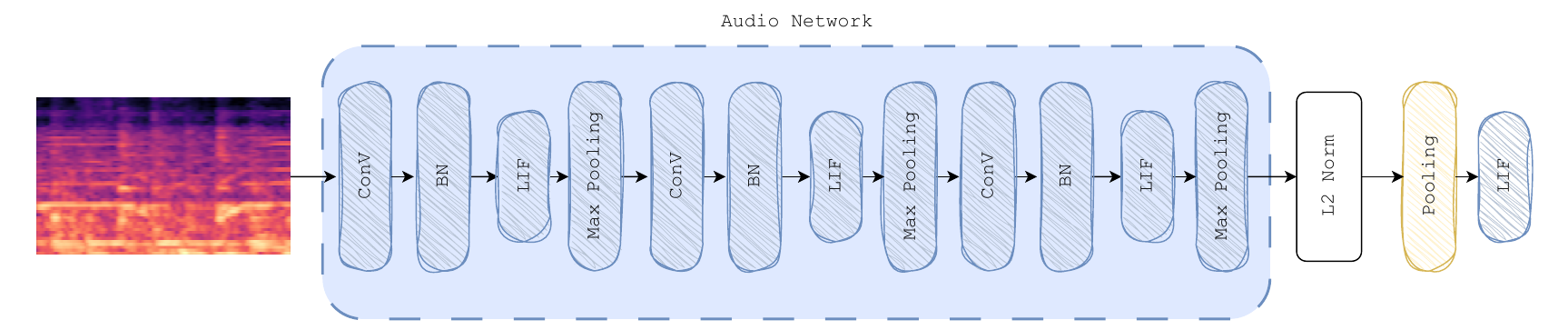}
    \caption{The proposed spiking audio feature extraction network, with the L2 normalization layer for audio feature discrimination. }
    \label{fig:audionet}
\end{figure*}

The last layers in the state-of-the-art spiking ResNet approaches consist of a fully connected layer responsible for the final classification task.
Here, we implement an L2 normalization layer as described in Section~\ref{sec:norm}, to ensure that the visual features lie on a hypersphere, thus enabling enhanced feature discrimination capabilities. 
Therefore, the L2 normalization layer generates the normalized feature vectors $V\in\mathbb{R}^{512}$.
The logits are then passed through the last LIF layer.
In contrast to other approaches in the field, the output layer of our L2-ActSpikeNett does not rely on the cross-entropy loss.
The output LIF layer generates spike-based predictions, producing spikes corresponding to class logits.
In particular, the total number of spikes for the $i$-th class is computed as:
\begin{equation}
S_i =  \sum^{T}_{t=1}{s_{i,t}},
\end{equation}
where $S_i$ the total number of spikes for class $i$, $s_{i,t}$ the spike output of the $i$-th neuron at timestep $t$, and $T$ the total number of timesteps. 
The target number of spikes $C_i$ is calculated as follows:
\begin{equation}
C_i = 
    \begin{cases}
    T\cdot r_{correct}, \quad\text{if class i is the correct class,} \\
    T\cdot r_{incorrect}\quad  \text{if class i is the incorrect class}.
    \end{cases}
\end{equation}
where $r_{correct}$ is the firing frequency of the correct class as a ratio, which defines how often the correct class neuron should fire relative to the total timesteps $T$, and $r_{incorrect}$ is the firing frequency of incorrect classes as a ratio.
The network is trained to minimize the mean squared error between the predicted and the target number of spikes along the number of timesteps, as follows:
\begin{equation}
\pazocal{L}_{MSE} = \frac{1}{B} \sum^{B}_{i=1}\frac{1}{N} \sum^{N}_{j=1}(S_{i,j}-S^{target}_{i,j})^2,
\end{equation}
where $B$ the batch size, $N$ the number of output classes, $S_{i,j}$ the total number of spikes for the $j$-th class of the $i$-th sample and $S^{target}_{i,j}$ the target number of spikes for the $j$-th class of the $i$-th sample.
The target spike count $C_i$ is used for the target matrix $S^{target}$, which specifies the desired number of spikes for each class. 
For the correct class $i$, the target is set to $C_i = T \cdot r_{correct}$, while for incorrect classes, it is set to $C_i = T \cdot R_{incorrect}$. 
The loss function minimizes the mean squared error between the predicted spike counts $S_{i,j}$ and these target values $S^{target}_{i,j}$.



\subsection{Audio Features}\label{sec:audioft}
In contrast to visual modality, few spiking architectures have been implemented for audio data.
To extract meaningful representations from the audio modality we initially pre-processed the audio datasets to produce the well-established spectrogram representation, following the methodology described in ~\cite{10293172}. 
According to that, the filterbank representations are generated by applying a short-time Fourier transformation (STFT) to the waveforms.
The resulting spectrograms are mapped to mel-spectrograms with 64-mel frequency bins and a maximum frequency of $8000$ Hz and then transformed to log-mel spectrograms using the power-to-db conversion.
We process the log-mel spectrograms in our audio feature network as shown in Fig.~\ref{fig:audionet}.
The audio feature network, consisting of convolutional, BN, Max Pooling and LIF neurons, constitutes a simpler compared to the visual network model mainly due to the simpler structure of the audio data.
Similarly to the visual feature network, the audio features are generated by the L2 normalization layer described in section~\ref{sec:norm}.
Following the normalization layer, the last LIF layer is used to produce spikes.
For the classification, we deploy the mean square error spike count loss, as in the visual feature network.
After training the audio network, the audio feature space $A\in\mathbb{R}^{27136}$ is used as an input along with the visual feature space to the feature fusion network.

\subsection{Features Fusion}\label{sec:fusion}
\begin{figure}
    \centering
    \includegraphics[scale=0.6]{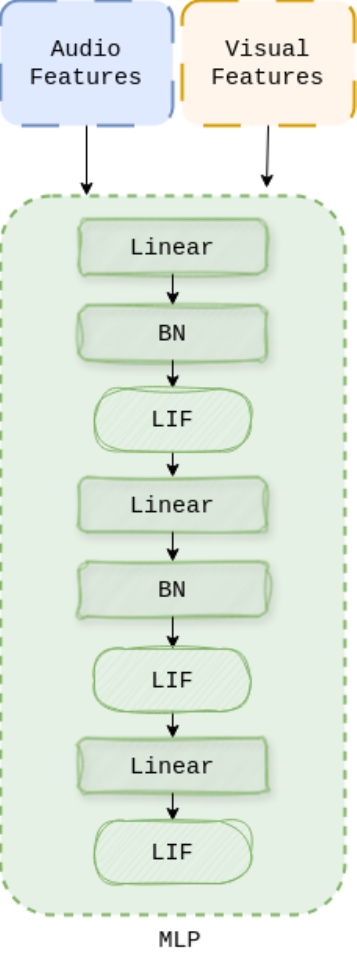}
    \caption{Architecture of the Spiking Multilayer Perceptron (SMLP) for multi-modal fusion, integrating normalized visual and audio feature embeddings for classification. }
    \label{fig:mlp}
\end{figure}

The temporal dynamics are leveraged across both modalities, with spiking neurons ensuring bio-plausible processing and efficient temporal feature extraction. 
Aiming to fuse the extracted feature vectors, we designed our SMPL approach as shown in Fig.~\ref{fig:mlp}.
The extracted embeddings from both networks are fed into a common latent space using linear projection layers, ensuring dimensional alignment and effective modality fusion.
Thus, we concatenated the features from both modalities. 
The visual features $V\in\mathbb{R}^{512}$ along with the auditory features $A\in\mathbb{R}^{27136}$ were presented as a unified representation in the SMLP as $V_{concat}\in\mathbb{R}^{27648}$.
After the concatenation, we apply batch normalization to smooth the distributions.
The smoothed features are then trained through a network with LIF, linear and batch normalization layers.
Similarly to the audio-visual networks, the MLP is also trained using the mean square error spike count loss.

We train our datasets with the SMLP for 10 epochs, with Adam optimizer, a 32 batch size and 8 timesteps.

\section{Experiments}
This section thoroughly evaluates our approach and compares it against other state-of-the-art techniques. 
Firstly, we present the employed datasets and learning strategies. 
Then, we evaluate the results to gain valuable insights into the behaviour of multi-modal networks during training.
Lastly, we perform an ablation study to support our architecture and findings.

\subsection{Experimental Setup}
Our models were implemented using the snnTorch and PyTorch~\cite{paszke2017automatic} libraries, written in Python.
Training and evaluation have been conducted on an NVIDIA GeForce RTX 3090 GPU with a total memory of 24GB.

We employed a CUDA-enabled environment to leverage GPU acceleration for faster matrix operations and neural network computations. 
The training pipeline was configured for multi-step temporal processing, where both visual and audio modalities were processed under a fixed number of timesteps.

\subsection{Datasets' Preparation}
To evaluate the performance of our approach, we conducted experiments on two audio-visual datasets, \textit{viz.}, CIFAR10-AV and UrbanSound8k-AV~\cite{10293172}.
In this section, we describe the datasets utilized in our approach along with the configuration parameters deployed for each dataset and modality.

\subsubsection{CIFAR10-AV}\label{subsec:cifar-dat}
The CIFAR10-AV~\cite{10293172} dataset extends the well-established CIFAR-10 dataset by introducing an auditory modality, thus producing an extended set of multi-modal audio-visual samples.
Each sample includes a common 32x32 pixel color image from the CIFAR-10 dataset and a $5$-second audio clip that corresponds to the given image category. 
The audio-visual CIFAR10-AV dataset contains a total of 60000 samples across the 10 classes.

For the visual modality, we normalized the image data in the interval $[0,1]$ and trained using a batch size of $128$ and $8$ timesteps.
Furthermore, we used the Adam optimizer~\cite{kingma2014adam} with an initial learning rate of $20^{-3}$ and a weight decay at $10^{-4}$.
The learning rate was adjusted with the MultiStepLR scheduler, using milestones at epochs $150$ and $275$ over a total of $300$ training epochs.
We employed the MSE count loss, setting the correct rate to $0.90$ and the incorrect rate to $0.10$.
Additionally, for gradient differentiation, we employed the surrogate gradient of the Heaviside step function~\cite{fang2021incorporating}, which approximates the gradient using a shifted arctan function. 
The forward pass is defined as follows:

\begin{equation}\label{eq:atan}
S = 
    \begin{cases}
    1,\quad \text{if $V\geq V_{th}$,} \\
    0, \quad \text{if $V<V_{th}$}
    \end{cases}
\end{equation}
where $V$ represents the membrane potential and $V_{th}$ the firing threshold.
For the backward pass, we approximate the gradient using the arctan function:
\begin{equation}\label{eq:atan}
S\approx \frac{1}{\pi}arctan(\pi V\frac{a}{2})
\end{equation}
with the gradient defined as:
\begin{equation}
\frac{\partial S}{\partial V} = \frac{1}{\pi}\frac{1}{(1+(\pi V \frac{a}{2})^2)}
\end{equation}
where $a$ controls the sharpness of the arctan approximation.

The SNN for the audio modality is trained using a batch size of 32 and 4 timesteps.
The Adam optimizer was employed for optimization with an initial learning rate of $10^{-4}$.
We employed the MSE count loss, setting the correct rate to $0.85$ and the incorrect rate to $0.15$.
For backpropagation, we utilized a fast sigmoid function with a slope of $k=5$ to approximate the gradient of the spike.
The forward pass of the spiking mechanism is defined as:
\begin{equation}\label{eq:atan}
S = 
    \begin{cases}
    1,\quad \text{if $V\geq V_{th}$,} \\
    0, \quad \text{if $V<V_{th}$}
    \end{cases}
\end{equation}
where similarly $V$ represents the membrane potential and $V_{th}$ the firing threshold.

The backward pass is approximated using the fast sigmoid function:
\begin{equation}\label{eq:atan}
S\approx \frac{V}{1+k |V|}
\end{equation}
with the gradient given by:
\begin{equation}
\frac{\partial S}{\partial V} =  \frac{V}{(1+k |V|)^2}
\end{equation}
\begin{figure*}[!t]
    \centering
    \subfloat[L2-norm Train Visual Features]{%
        \includegraphics[width=0.25\textwidth]{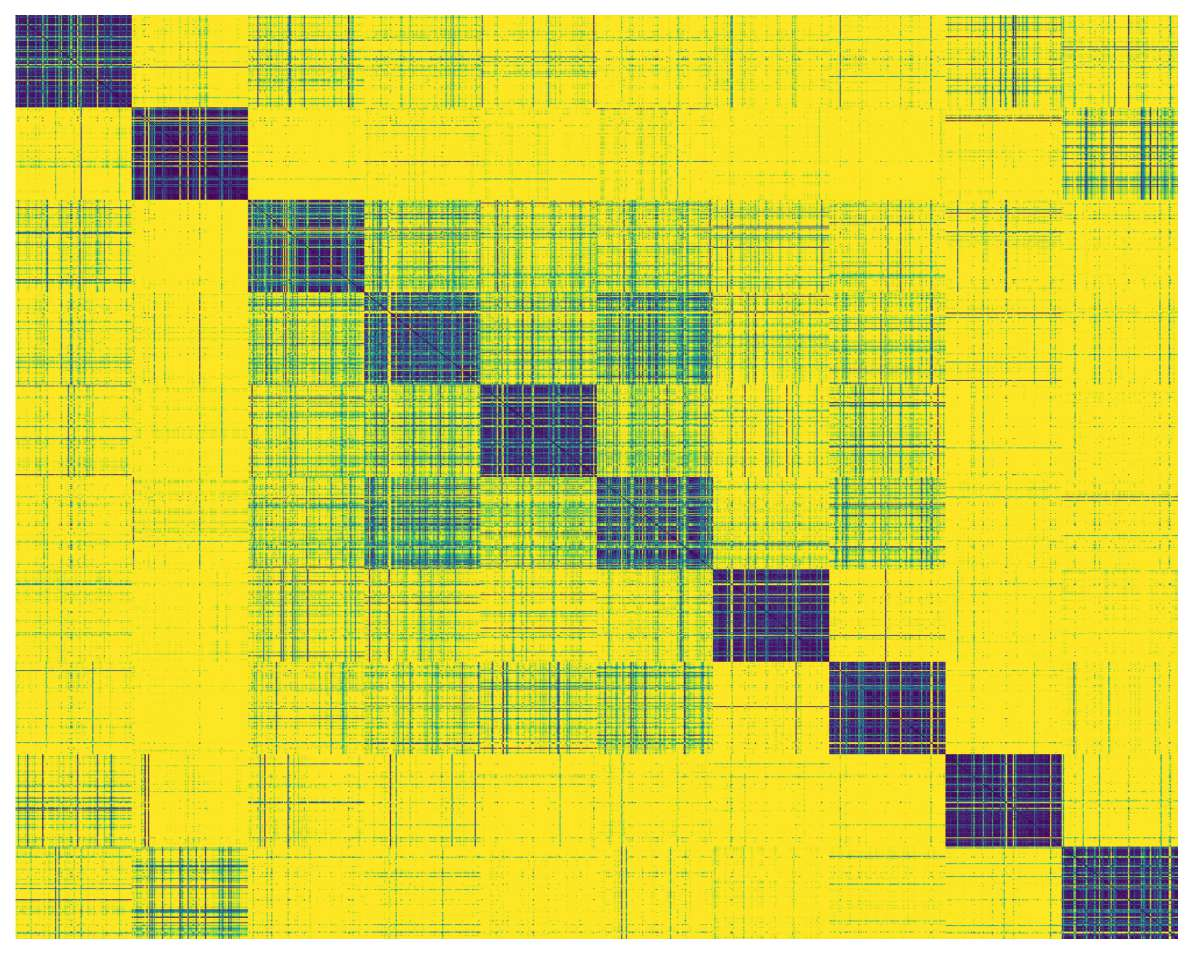}%
        \label{subfig:l2normtrainvisual_cif}}
    \hfill
    \subfloat[Vanilla Train Visual Features]{%
        \includegraphics[width=0.25\textwidth]{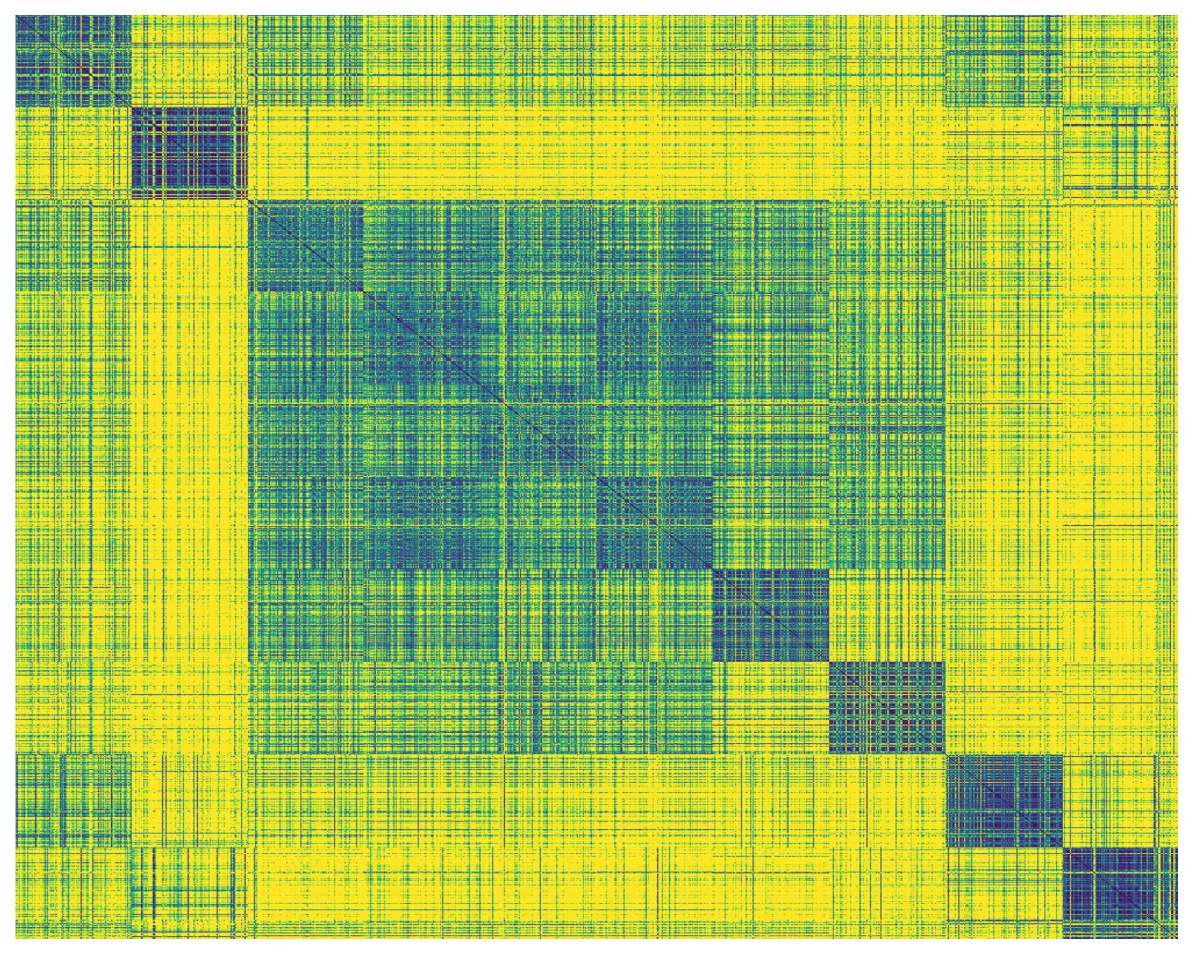}%
        \label{subfig:vaniltrainvisual_cif}}
    \hfill
    \subfloat[L2-norm Train Audio Features]{%
        \includegraphics[width=0.25\textwidth]{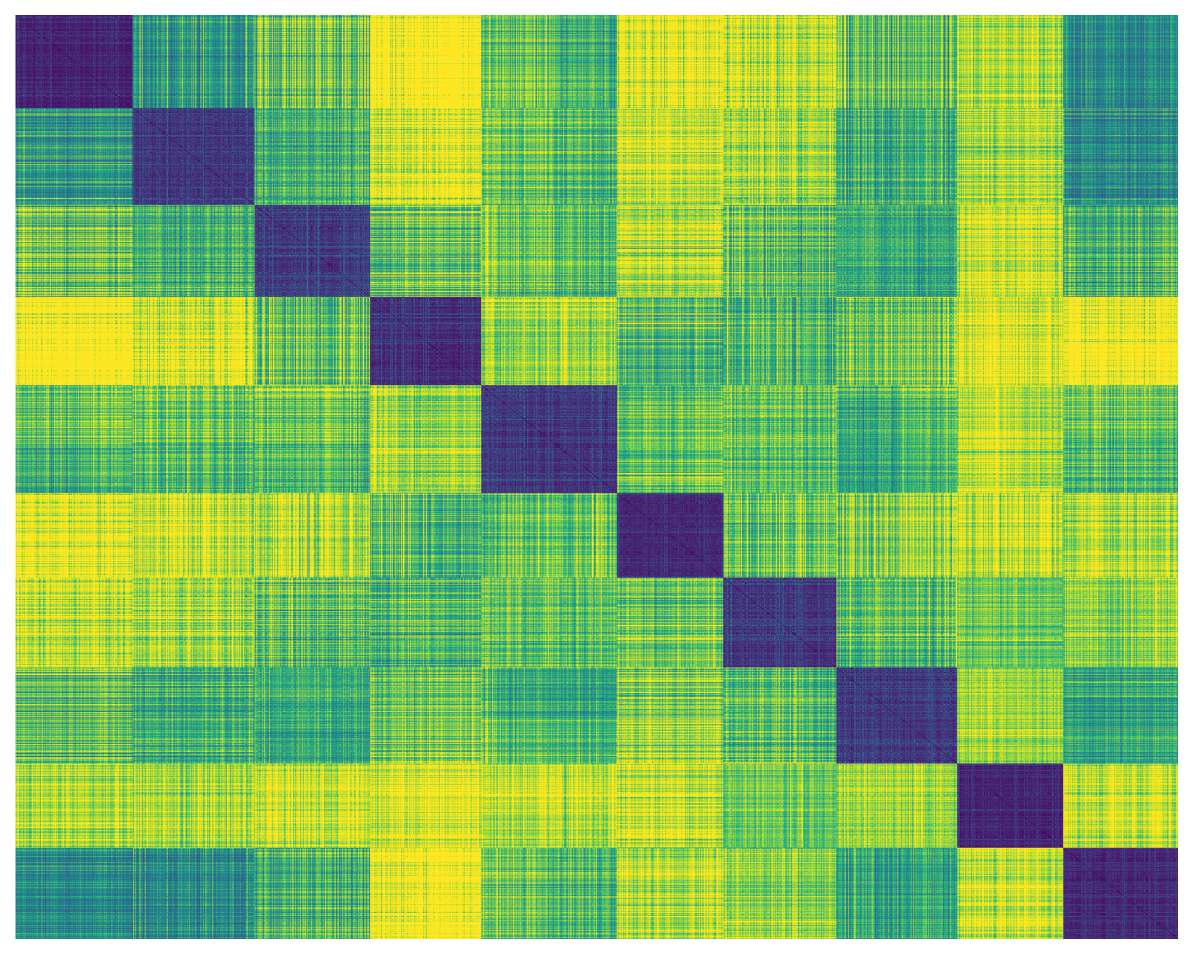}%
        \label{subfig:l2trainaudio_cif}}
    \hfill
    \subfloat[Vanilla Train Audio Features]{%
        \includegraphics[width=0.25\textwidth]{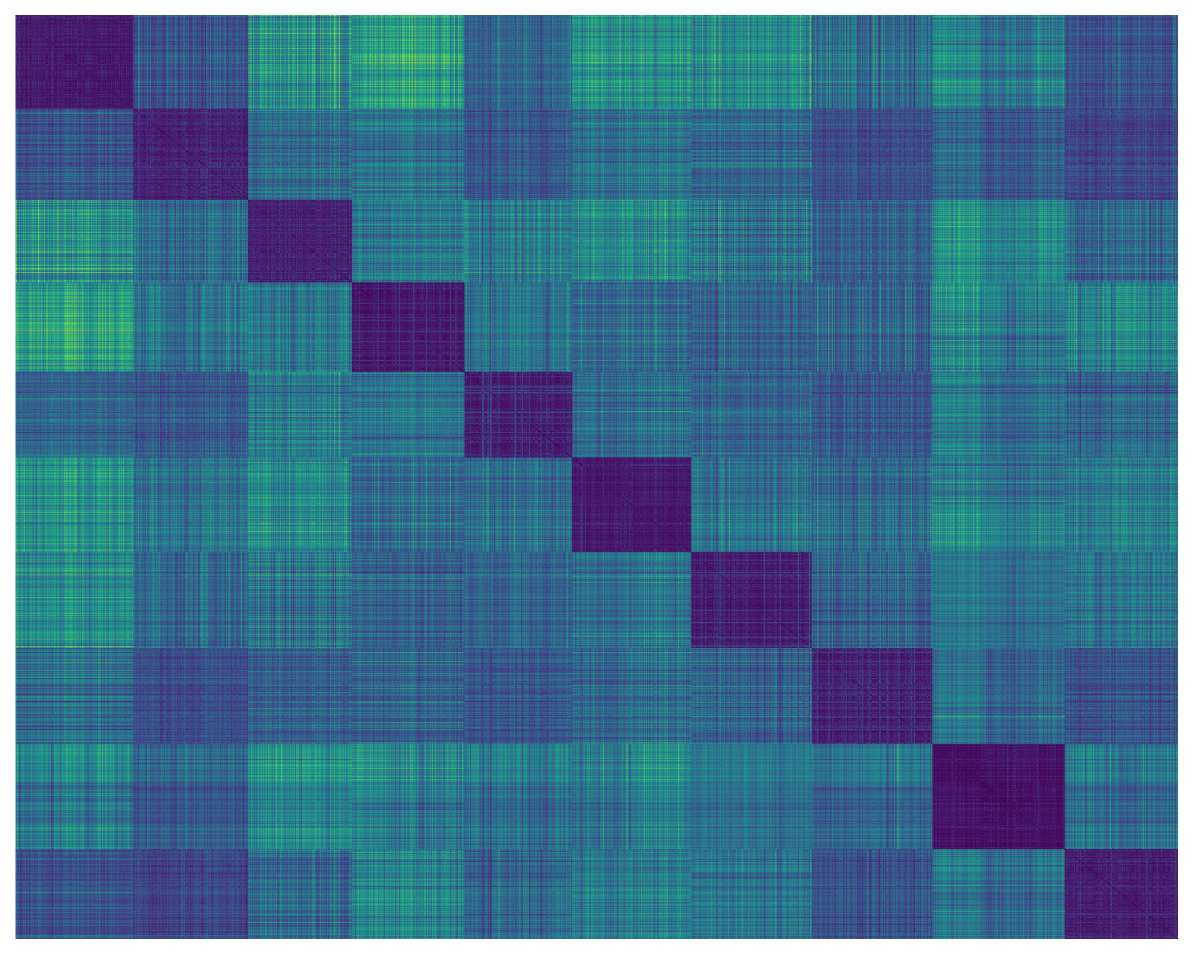}%
        \label{subfig:vaniltrainaudio_cif}}
    \par\bigskip
    \subfloat[L2-norm Test Visual Features]{%
        \includegraphics[width=0.25\textwidth]{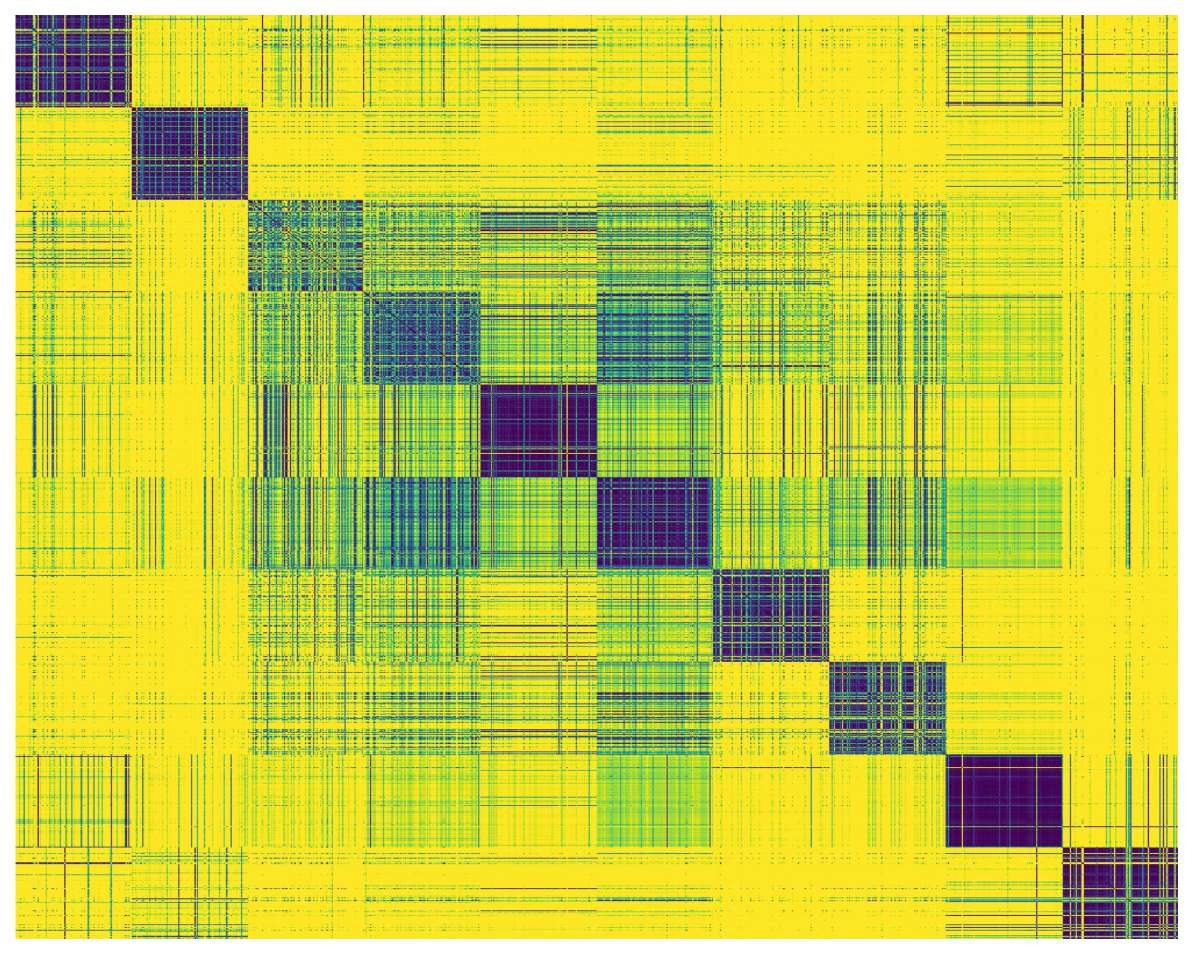}%
        \label{subfig:l2testvisual_cif}}
    \hfill
    \subfloat[Vanilla Test Visual Features]{%
        \includegraphics[width=0.25\textwidth]{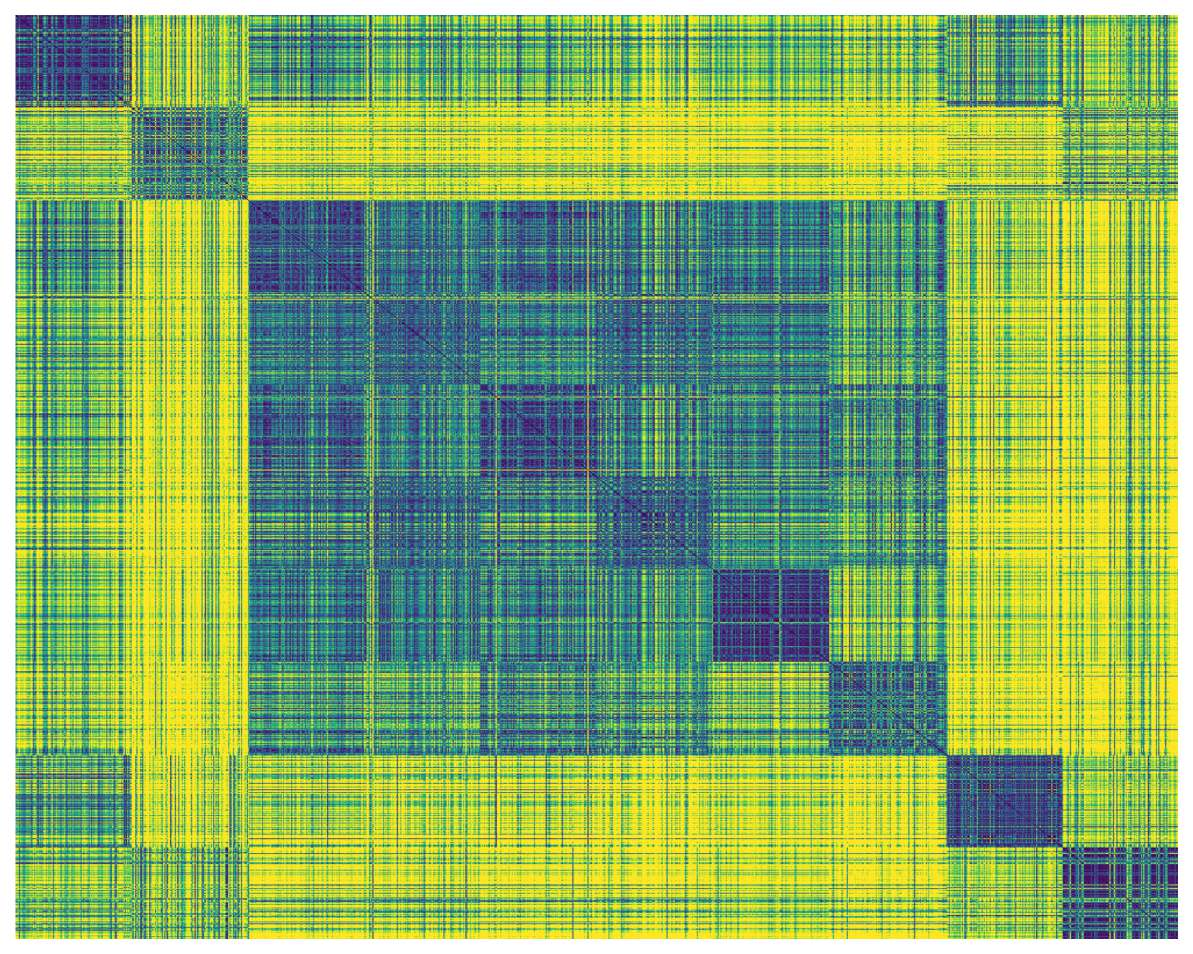}%
        \label{subfig:vaniltestvisual_cif}}
    \hfill
    \subfloat[L2-norm Test Audio Features]{%
        \includegraphics[width=0.25\textwidth]{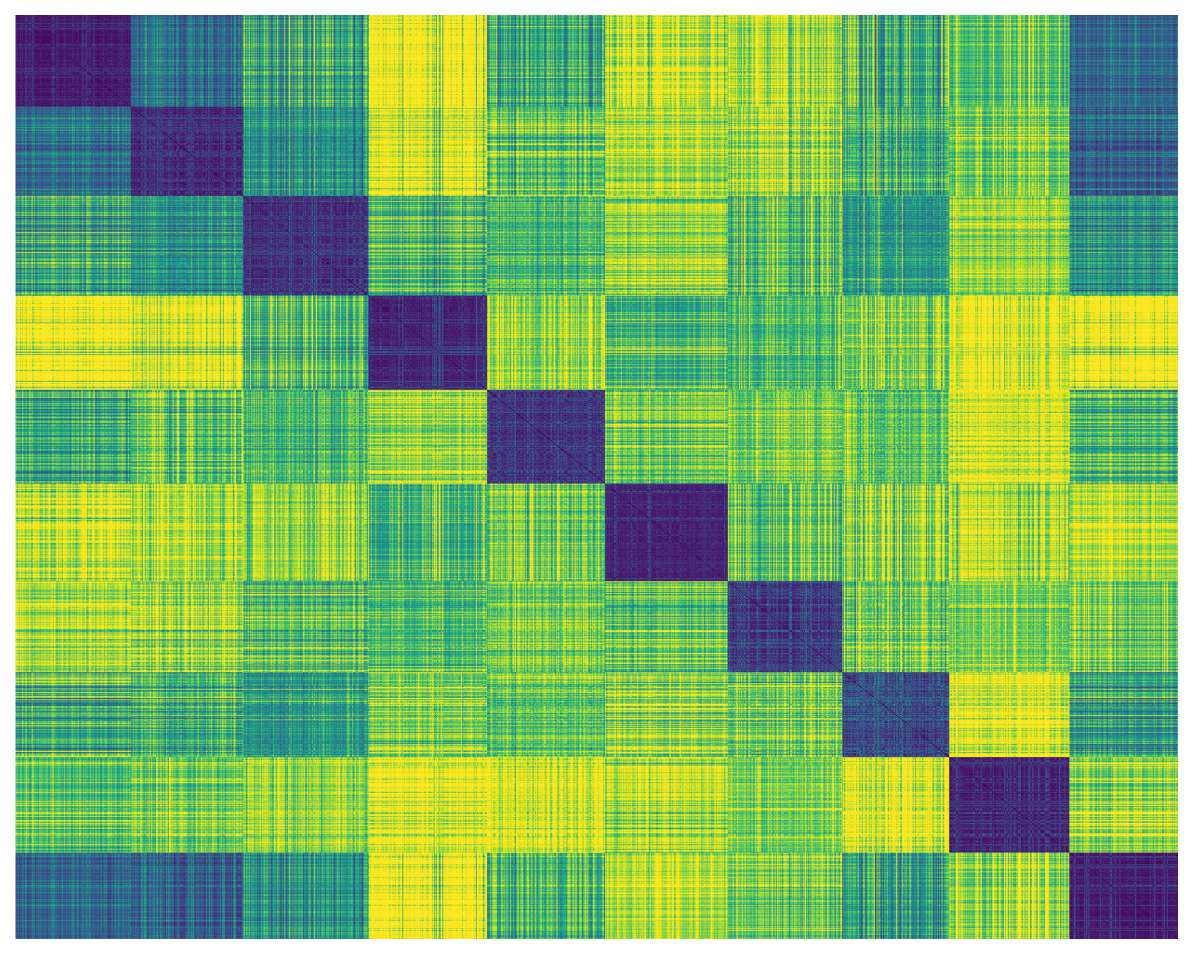}%
        \label{subfig:l2testaudio_cif}}
    \hfill
    \subfloat[Vanilla Test Audio Features]{%
        \includegraphics[width=0.25\textwidth]{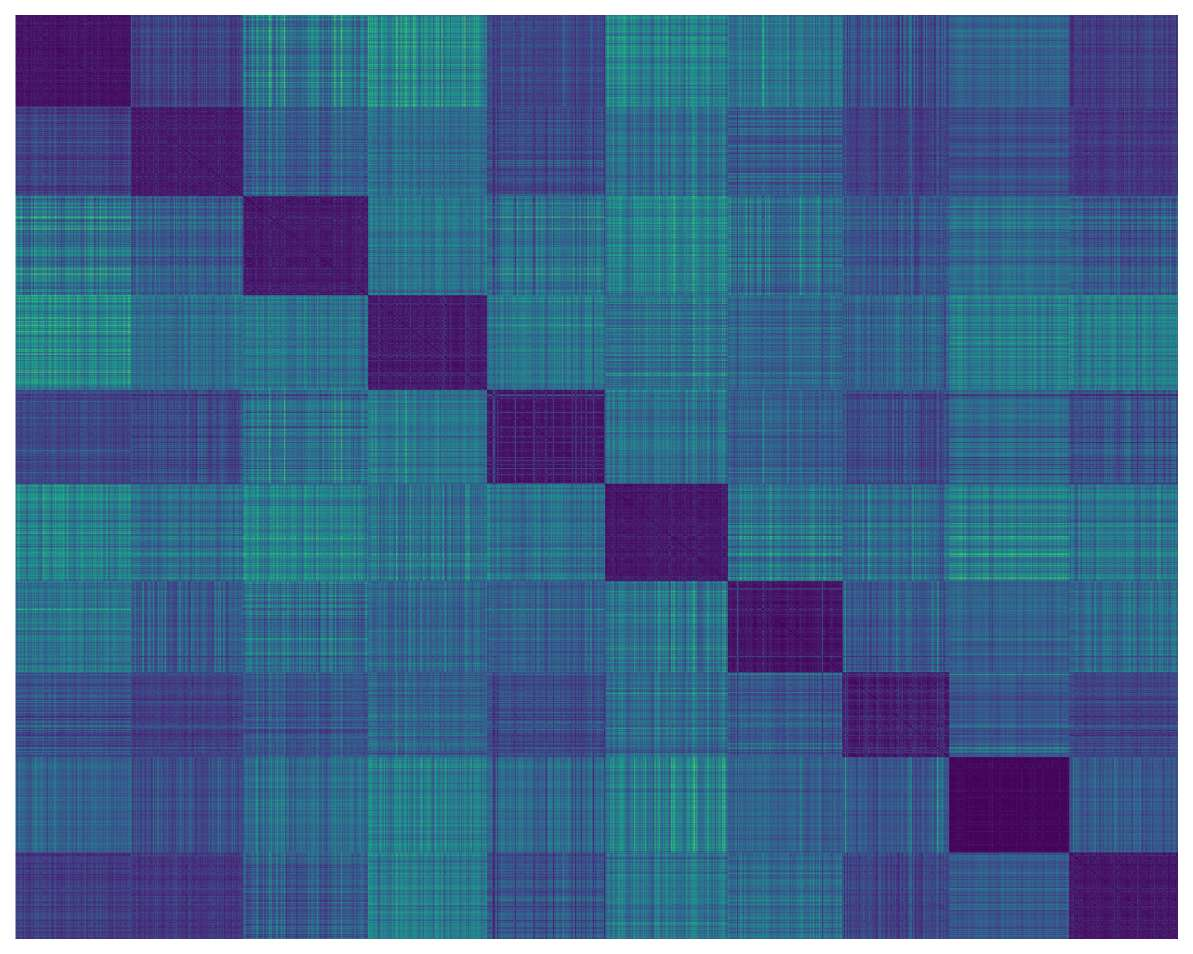}%
        \label{subfig:vanillatestaudio_cif}}
    \caption{Cosine distance matrices of CIFAR10-AV features comparing L2-normalized and vanilla models. Figs. ~\ref{subfig:l2normtrainvisual_cif},~\ref{subfig:l2trainaudio_cif},~\ref{subfig:l2testvisual_cif},~\ref{subfig:l2testaudio_cif}, show L2-normalized visual/audio features, demonstrating stronger intra-class compactness and inter-class separability. Figs.~\ref{subfig:vaniltrainvisual_cif},~\ref{subfig:vaniltrainaudio_cif},~\ref{subfig:vaniltestvisual_cif},~\ref{subfig:vanillatestaudio_cif}, depict features from a vanilla model with lower separability.}
    \label{fig:cifar10mat}
\end{figure*}
\subsubsection{UrbanSound8K-AV}
The UrbanSound8K-AV~\cite{10293172} dataset enhances the UrbanSound8K audio dataset by incorporating a visual modality.
Each sample includes a color image with square resolution, paired with a $4$-second audio clip that aligns with the respective class. 
The audio-visual UrbanSound8K-AV includes 8732 multi-modal samples covering ten sound categories.

For the visual modality, the normalized data has been trained using a batch size of 64 and 8 timesteps. The remaining parameters, including the optimizer, learning rate, and loss function, are identical to those used for the visual modality of the CIFAR10-AV dataset. Similarly, the training setup for the audio modality follows the same configuration as described for the audio modality in the CIFAR10-AV dataset.

\begin{table}
    \caption{Comparison of Modalities on CIFAR10-AV}
    \vspace{5pt} 
    \centering
    \setlength{\tabcolsep}{6pt} 
    \renewcommand{\arraystretch}{1.2} 
    \begin{tabular}{lc} 
        \hline
        \textbf{Modality} & \textbf{Accuracy (\%)} \\
        \hline
        Image Modality & 92.74 \\
        Audio Modality & 99.60 \\
        Multi-modal Fusion & 98.60 \\
        \hline
    \end{tabular}
    \label{tab:cifarmod}
\end{table}

\subsection{Experimental results on CIFAR10-AV}

To assess our suggested model's performance, we ran several tests using the CIFAR10-AV dataset.
Table~\ref{tab:cifarmod} shows that our audio model performs exceptionally well at $99.6\%$, while our visual model attains an accuracy of $92.74\%$. 
Furthermore, the fusion achieves an overall accuracy of $98.6\%$.

Although these results demonstrate the efficiency of our model in the CIFAR10-AV dataset, it is equally important to highlight the high feature separability between different classes introduced by our approach for the first time in spiking classification.
Fig.~\ref{fig:cifar10mat} visually demonstrates the enhanced separation achieved by the proposed L2 normalization layer compared to the vanilla FC layer usually employed in SNNs.
For a complete overview, each figure illustrates all the cosine distances between each pair of accumulated feature vectors from the dataset for a given set, training or testing, and modality.

To achieve that, we calculated the cosine distances between two feature vectors, by applying the cosine distance formula:

\begin{equation}
C_{dist} = 1 - \frac{\textbf{f}_i \cdot \textbf{f}_j}{\lVert \textbf{f}_i \rVert \lVert \textbf{f}_j \rVert}
\end{equation}
where, $\textbf{f}_i$ and $\textbf{f}_j$ are the accumulated feature vectors among the network's timesteps for the samples $i$ and $j$, respectively.
Since samples from the same class are closely aligned in the feature space, this metric indicates that the diagonal members of the similarity matrix present strong intra-class similarity. 
On the other hand, off-diagonal elements demonstrate inter-class separability, where stronger class separation is represented by lighter colors.
A heatmap representing the resulting similarity matrix is displayed, with yellow denoting high angular distance and darker ones denoting darker ones.

\begin{figure}
    \centering
    \includegraphics[scale=0.4]{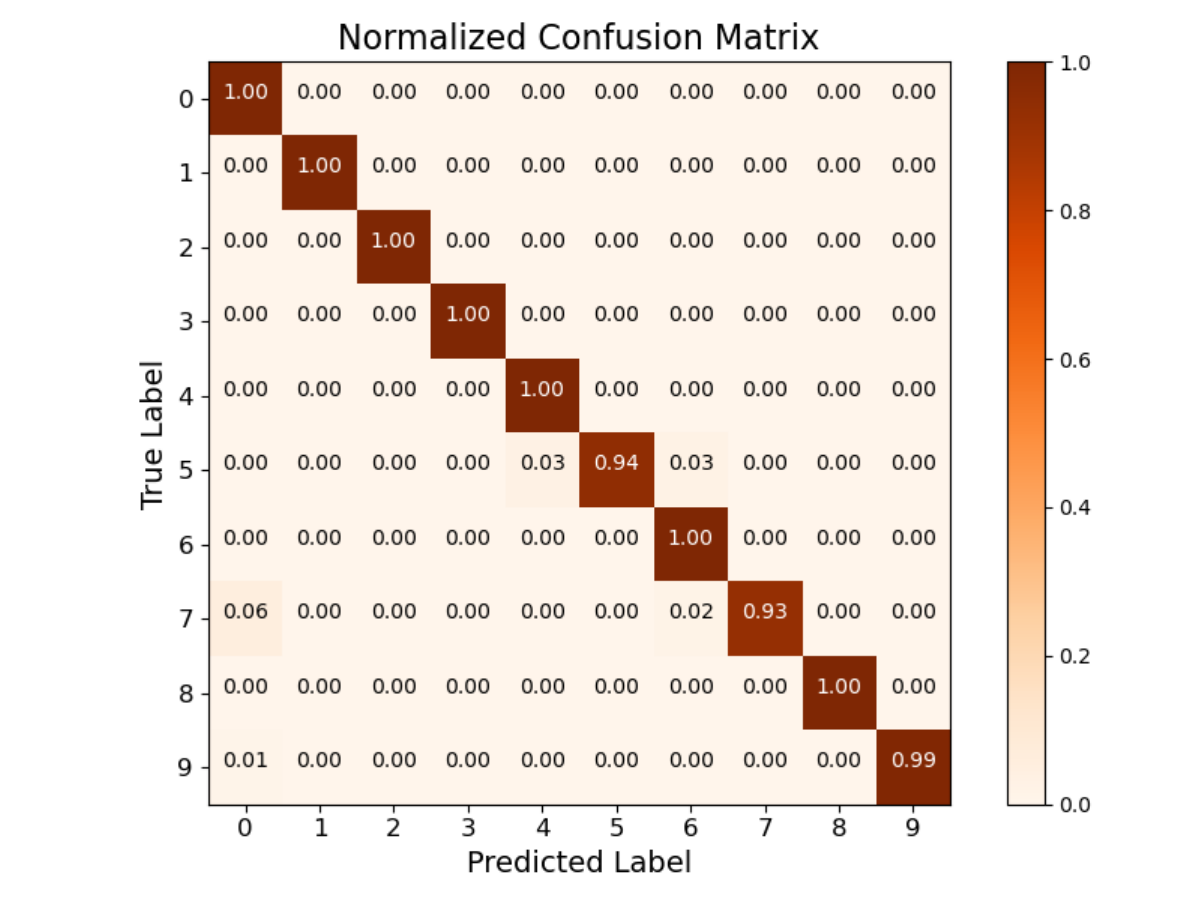}
    \caption{The normalized confusion matrix of the audio-visual model on the CIFAR10-AV dataset.}
    \label{fig:cifarconfmat}
\end{figure}

The outcomes of the training set with the suggested L2 normalization layer for the visual network are shown in Fig.~\ref{subfig:l2normtrainvisual_cif}.
Improved class separability is evident by observing the diagonal patterns
In contrast, the lighter colors illustrate the high angular distance between the accumulated feature vectors of different classes.
This is compared against Fig.~\ref{subfig:vaniltrainvisual_cif}, which depicts the results obtained by the accumulated feature vectors of a common FC layer in the SNN, denoted as vanilla SNN.
Similar findings can be concluded from the comparison between the distance matrices obtained from the visual feature vectors of the testing set in Figs.~\ref{subfig:vaniltestvisual_cif} and ~\ref{subfig:l2testvisual_cif}, thus ensuring that class separability remains in the learned weights of the network after training.

Subsequently, the same conclusions can be drawn by observing the accumulated feature vectors of the audio network.
More specifically, in Figs.~\ref{subfig:l2trainaudio_cif} and ~\ref{subfig:l2testaudio_cif} we can ascertain the supremacy of the L2 normalization layer against the common FC one depicted in Figs.~\ref{subfig:vaniltrainaudio_cif} and~\ref{subfig:vanillatestaudio_cif}, correspondingly, in terms of feature discrimination capacities.


In addition, we analyzed the per-class performance of the fusion model estimating the normalized confusion matrix, as shown in Fig.~\ref{fig:cifarconfmat}. 
The confusion matrix presents a thorough analysis of the classification performance among the ten classes.
The reader can discern that the majority of the classes achieve nearly perfect classification accuracy, demonstrating the outstanding performance of the fusion model.
In addition, there are only few off-diagonal non-zero values, indicating low misclassification rates.
The above fact highly demonstrates the efficient fusion of the feature vectors after the L2 normalization layer from both the auditory and visual modalities using SNNs for the final classification.

\subsection{Experimental results on UrbanSound8k-AV}

We conducted a similar experimental study on the UrbanSound8K-AV dataset to further evaluate our model's performance on the classification task. 
The classification performance of the two uni-modal architectures as well as the fusion model are presented in Table~\ref{tab:urbanmod}, demonstrating the efficacy of our approach in combining visual and auditory data. 
In particular, the accuracy of the auditory modality is $87.49\%$, while the visual modality is $95.42\%$.
With a considerably enhanced accuracy of $97.2\%$, the multi-modal fusion technique appears to highly capitalize on the performance achieved by the corresponding uni-modal models.

\begin{figure*}[!t]
    \centering
    \subfloat[L2-norm Train Visual Features]{%
        \includegraphics[width=0.25\textwidth]{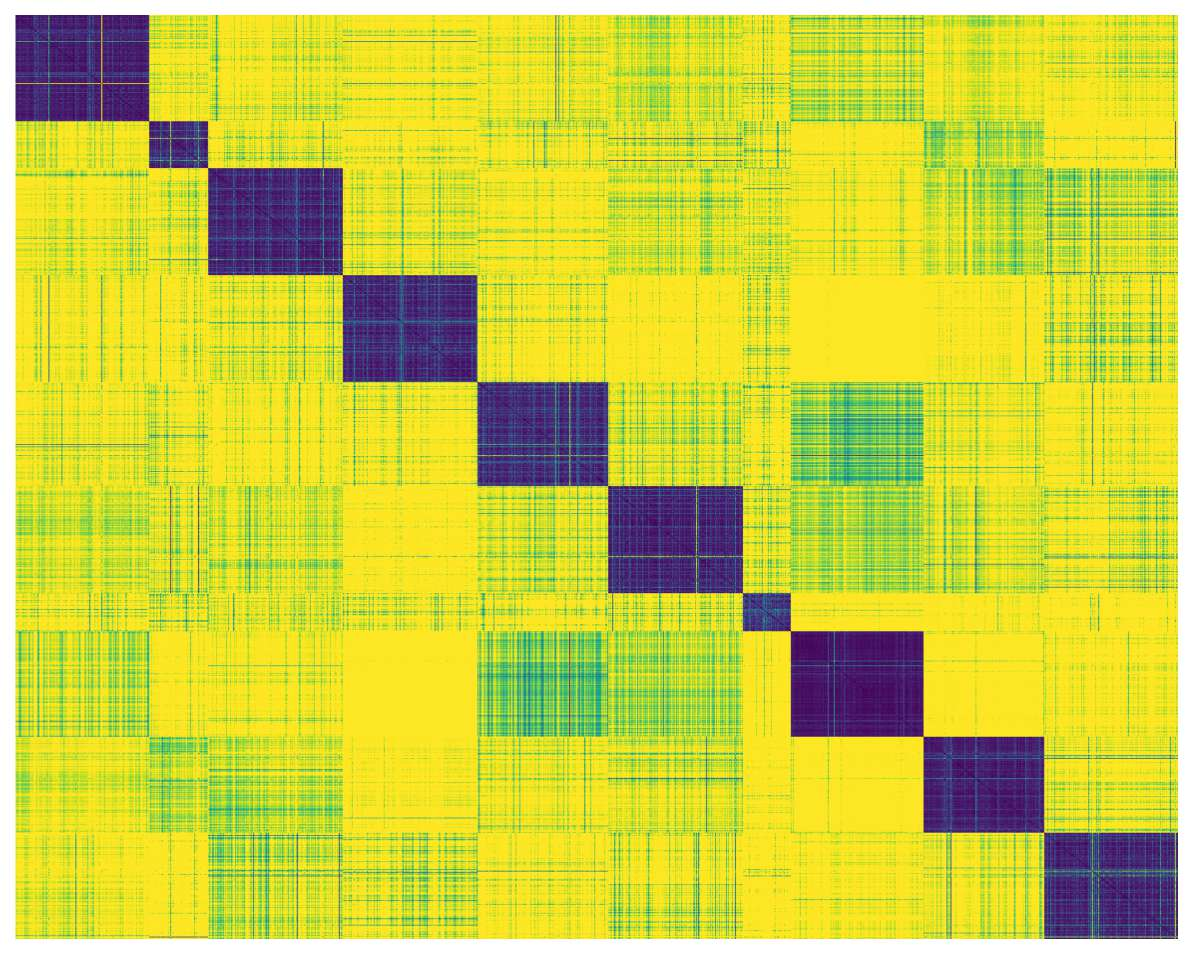}%
        \label{subfig:l2trainvisual}}
    \hfill
    \subfloat[Vanilla Train Visual Features]{%
        \includegraphics[width=0.25\textwidth]{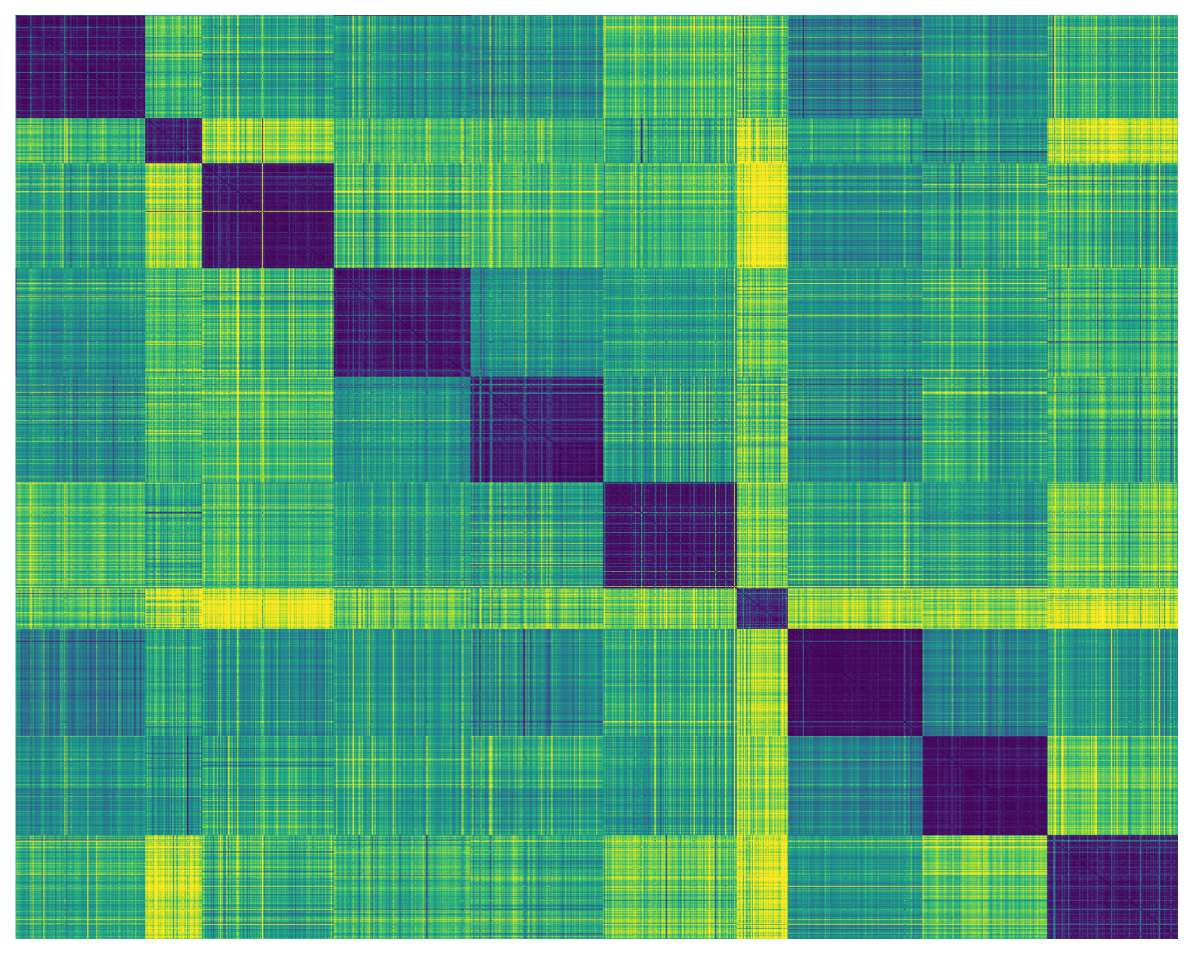}%
        \label{subfig:vaniltrainvisual}}
    \hfill
    \subfloat[L2-norm Train Audio Features]{%
        \includegraphics[width=0.25\textwidth]{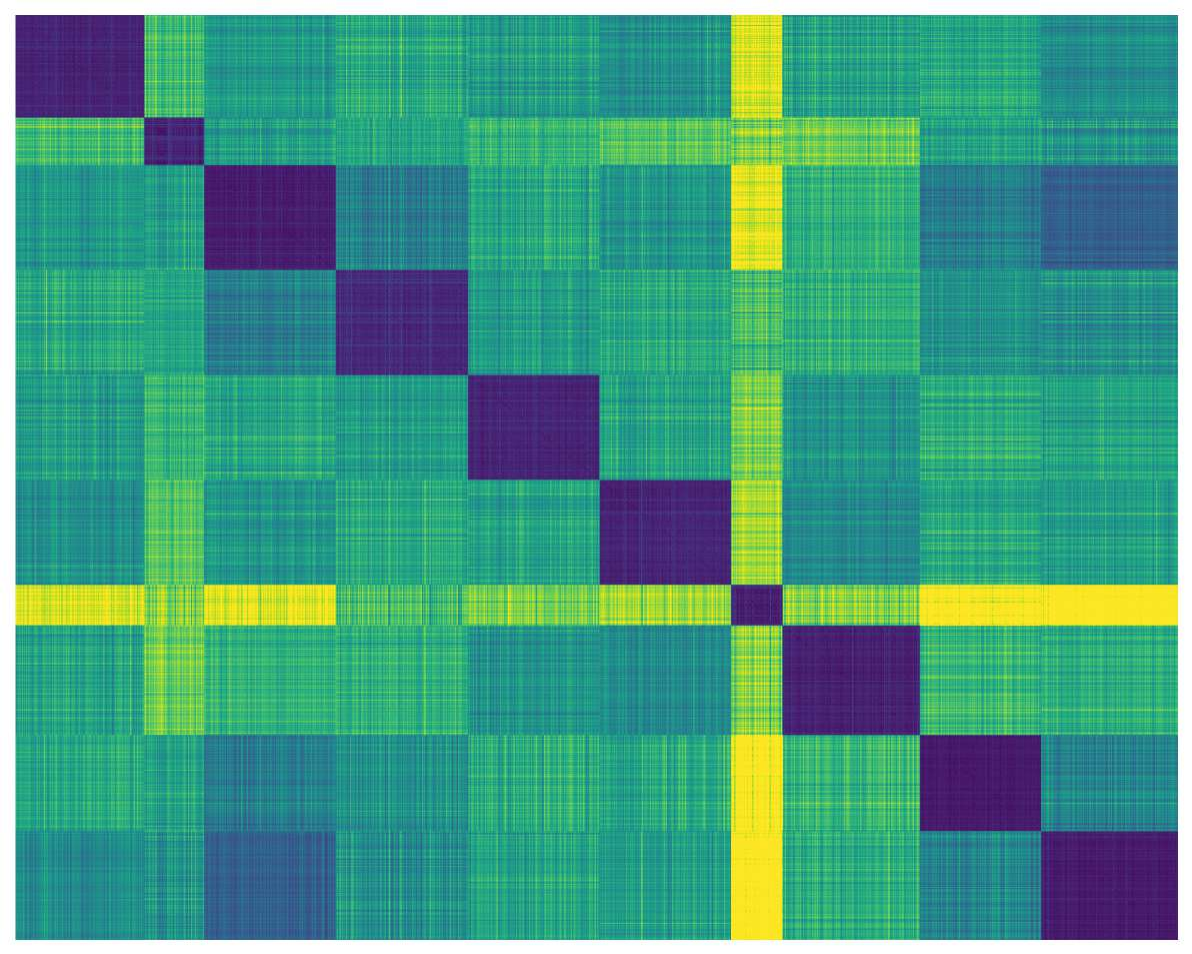}%
        \label{subfig:l2trainaudio}}
    \hfill
    \subfloat[Vanilla Train Audio Features]{%
        \includegraphics[width=0.25\textwidth]{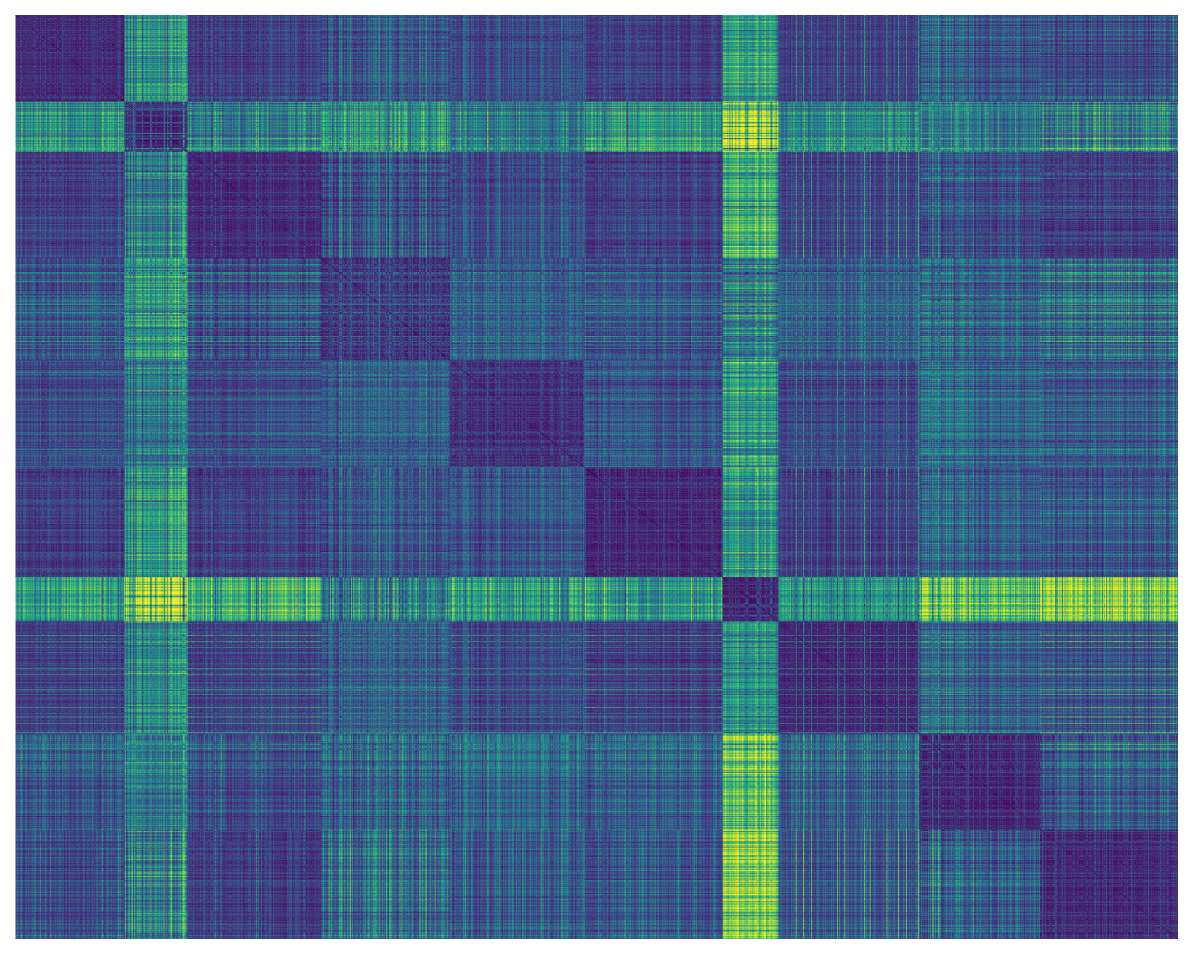}%
        \label{subfig:vaniltrainaudio}}
    \par\bigskip
    \subfloat[L2-norm Test Visual Features]{%
        \includegraphics[width=0.25\textwidth]{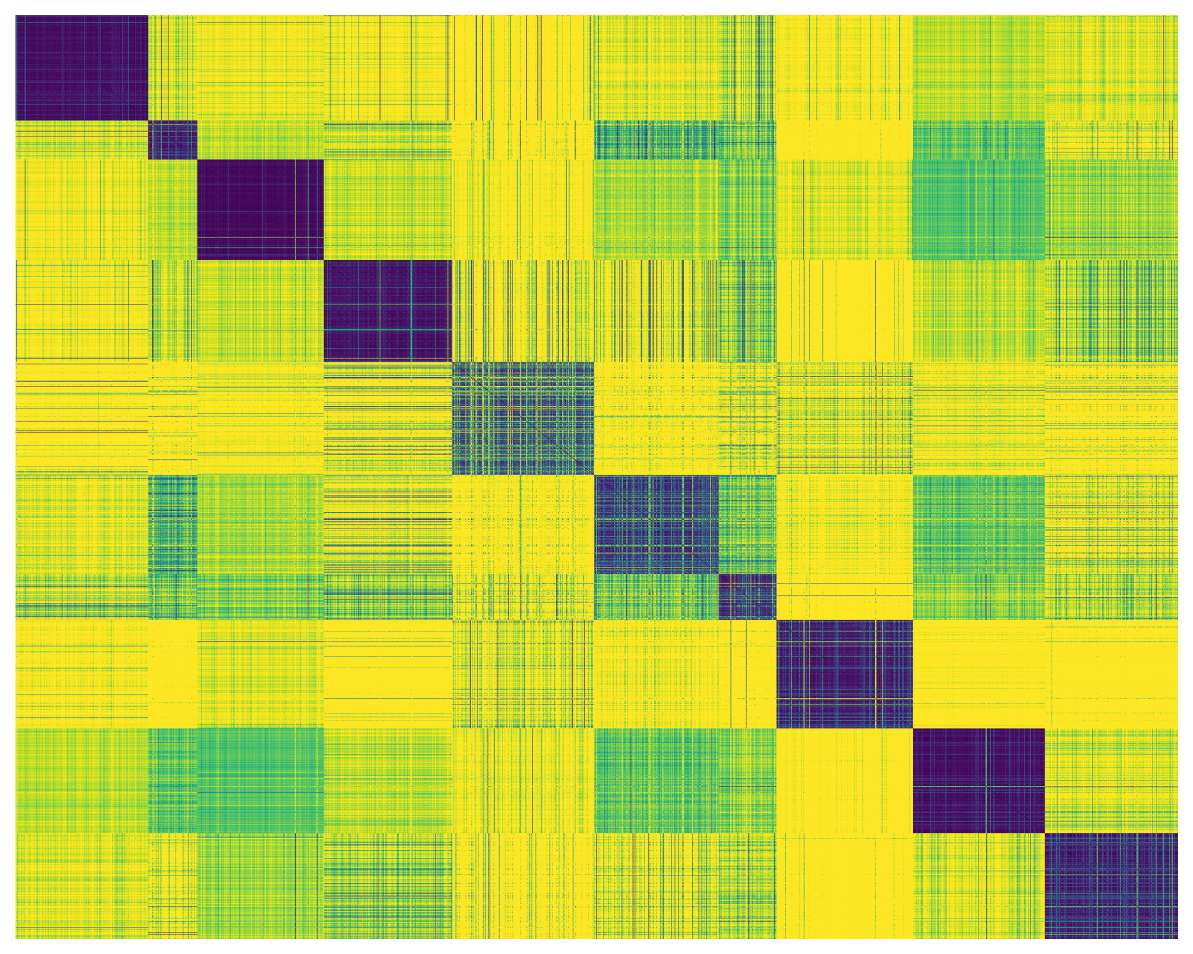}%
        \label{subfig:l2testvisual}}
    \hfill
    \subfloat[Vanilla Test Visual Features]{%
        \includegraphics[width=0.25\textwidth]{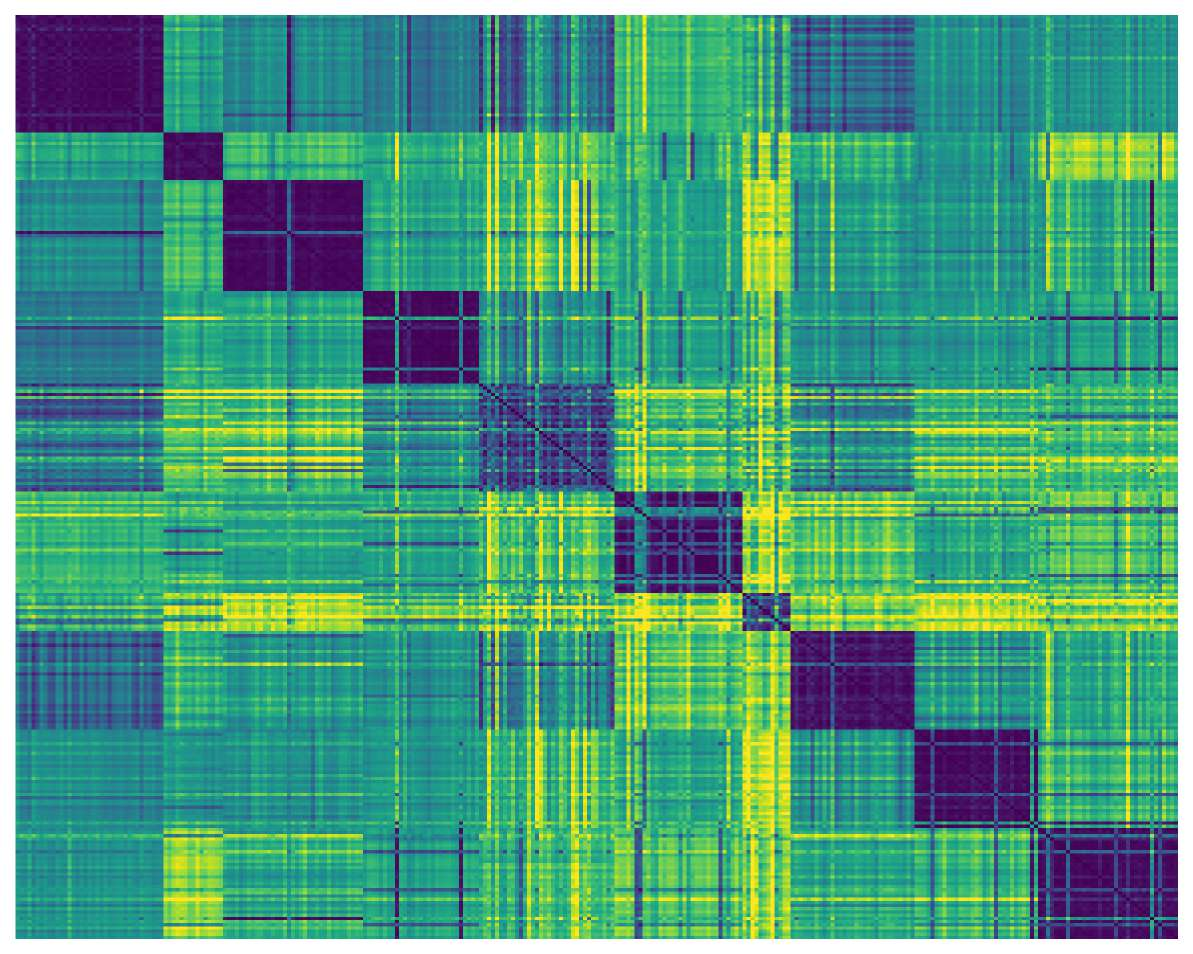}%
        \label{subfig:vaniltestvisual}}
    \hfill
    \subfloat[L2-norm Test Audio Features]{%
        \includegraphics[width=0.25\textwidth]{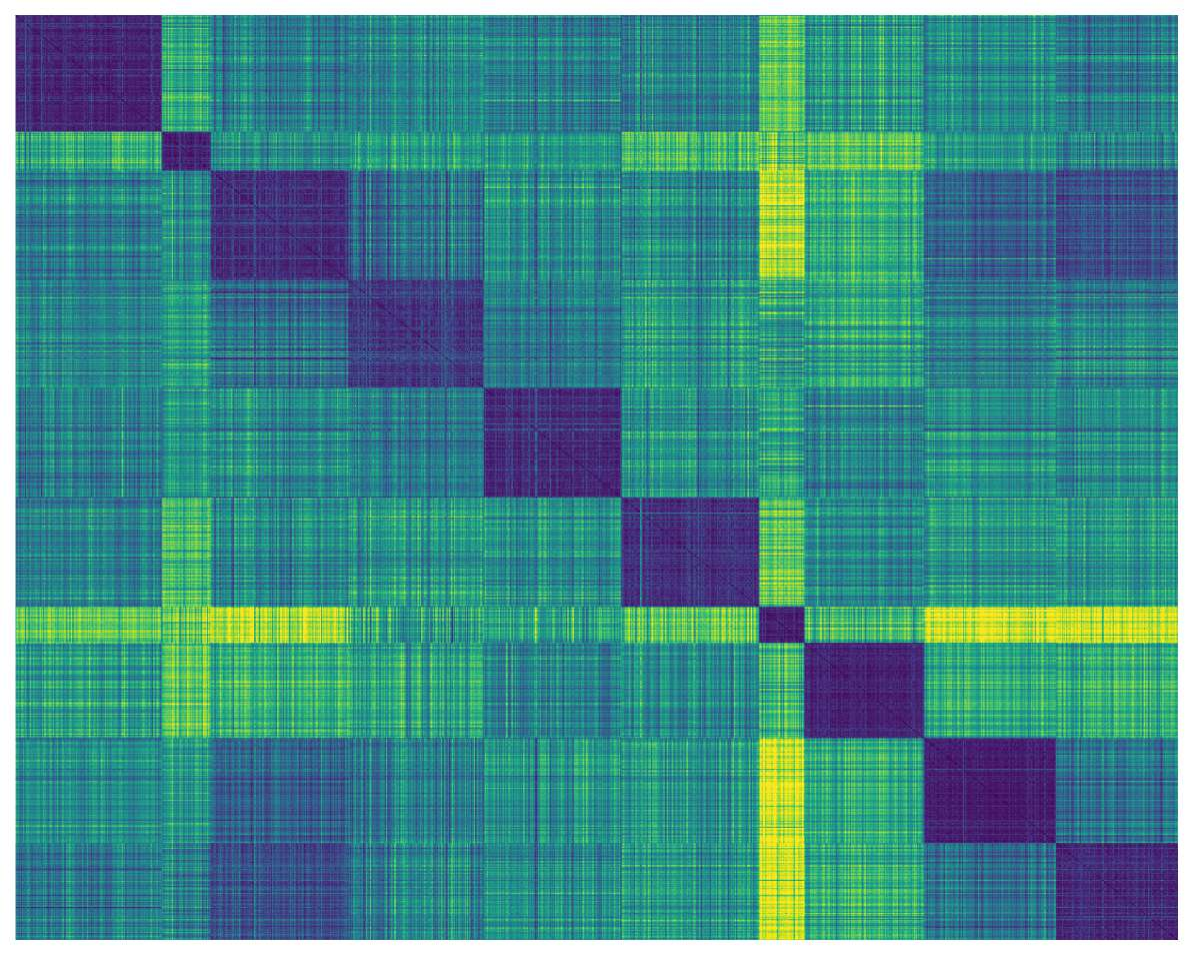}%
        \label{subfig:l2testaudio}}
    \hfill
    \subfloat[Vanilla Test Audio Features]{%
        \includegraphics[width=0.25\textwidth]{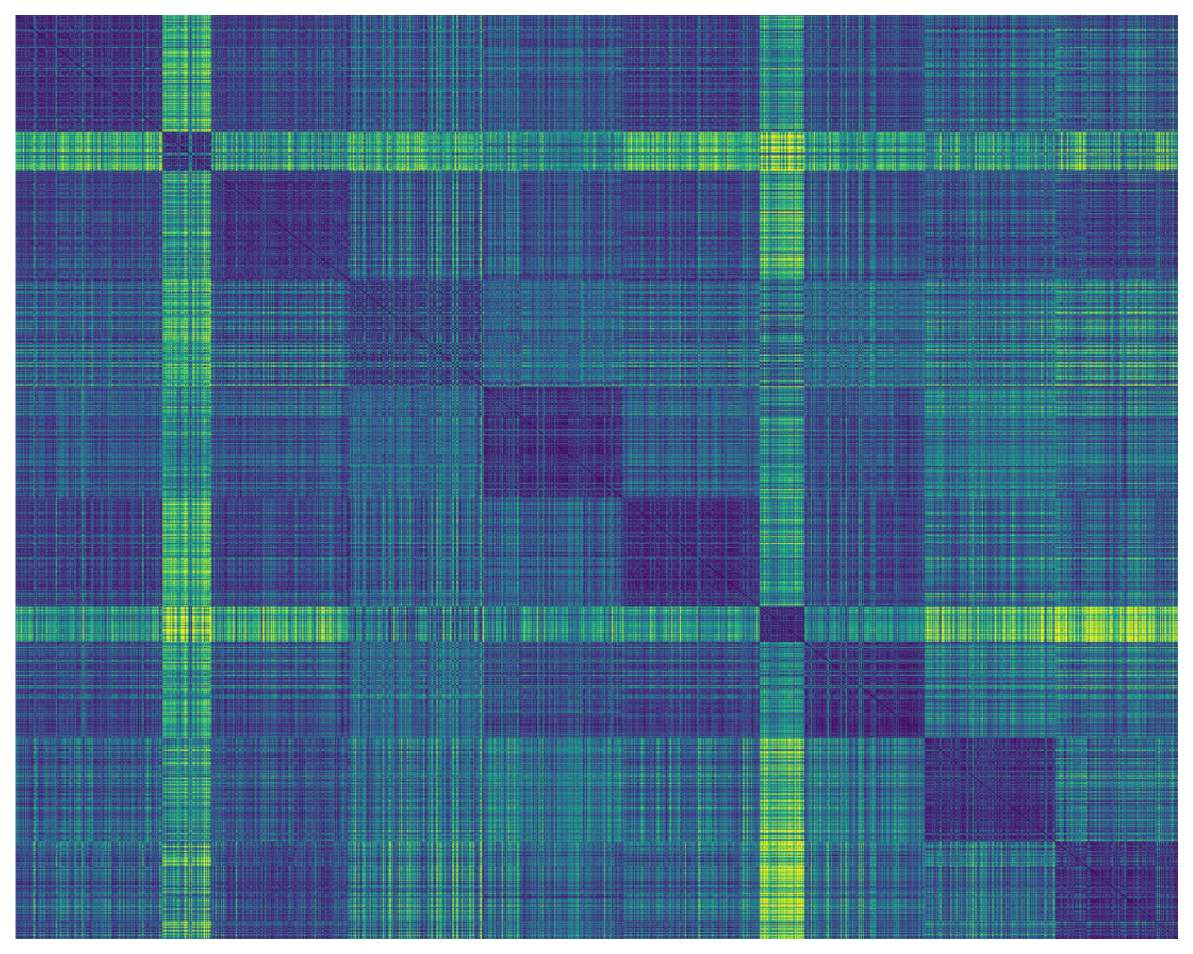}%
        \label{subfig:vaniltestaudio}}
    \caption{Cosine distance matrices of UrbanSound8K-AV features comparing L2-normalized and vanilla models.}
    \label{fig:urbanmat}
\end{figure*}

Moreover, in Fig.~\ref{fig:urbanmat} we illustrate the superiority of the L2 normalization layer against the common FC layer in feature discrimination.

\begin{table}[h]
    \centering
    \caption{Comparison of Modalities on UrbanSound8K-AV}
    \setlength{\tabcolsep}{6pt}
    \renewcommand{\arraystretch}{1.2}
    \begin{tabular}{lc}
        \hline
        \textbf{Modality} & \textbf{Accuracy (\%)} \\
        \hline
        Image Modality & 95.42 \\
        Audio Modality & 87.49 \\
        Multi-modal Fusion & 97.20 \\
        \hline
    \end{tabular}
    \label{tab:urbanmod}
\end{table}

\begin{figure}
    \centering
    \includegraphics[scale=0.4]{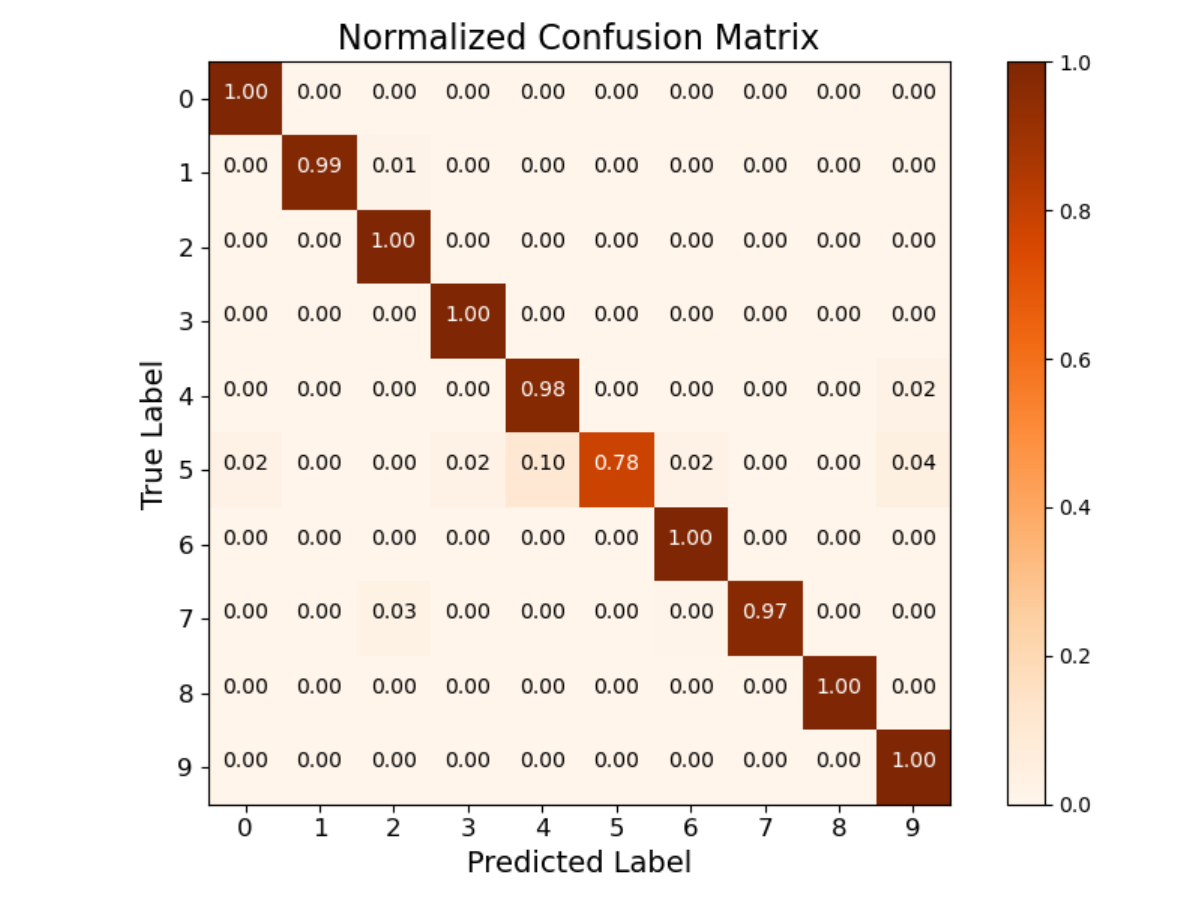}
    \caption{Normalized confusion matrix for the fusion model on UrbanSound8K-AV.}
    \label{fig:urbanconfmat}
\end{figure}

As demonstrated in Fig.~\ref{subfig:l2trainvisual} and Fig.~\ref{subfig:l2trainaudio} for the L2-normalized characteristics during training, considerable separability is indicated by the darker intensities in the diagonal elements of the distance matrix.
On the other hand, the vanilla network's Figs.~\ref{subfig:vaniltrainvisual} and \ref{subfig:vaniltrainaudio} show less clear discrimination between the diagonal elements and the rest ones, indicating most similarities across classes.
Additionally, Figs.~\ref{subfig:l2testvisual} and \ref{subfig:l2testaudio} show that the L2 normalization still exhibits higher separability on the testing dataset, in contrast to the one achieved by the vanilla network as depicted in Figs.~\ref{subfig:vaniltestvisual} and~\ref{subfig:vaniltestaudio}.
One should note that the dissimilar squares in the diagonal of the distance matrices are owed to the different sizes of the classes on the UrbanSound8k-AV dataset.
Finally, we utilized the normalized confusion matrix to evaluate the fusion model's classification performance for each class on the UrbanSound8K-AV dataset, as demonstrated in Fig.~\ref{fig:urbanconfmat}.
The significance of multi-modal integration can be realized through all classes' almost perfect classification outcomes, excluding class 5.
Hence, the reader can ascertain the approach's robustness and consistent behavior by considering the results on both CIFAR-AV and Urbansound-AV datasets.

\begin{table}[h]
    \centering
    \caption{Ablation Study Results}
    \setlength{\tabcolsep}{6pt}
    \renewcommand{\arraystretch}{1.2}
    \begin{tabular}{lcc}
        \hline
        \textbf{Modification} & \textbf{Modality} & \textbf{Accuracy (\%)} \\
        \hline
        LIF-after-BN & Visual & 90.40 \\
        LIF-before-Add & Visual & 86.82 \\
        Without Dropout & Audio & 78.16 \\
        Without 3rd Block & Audio & 72.36 \\
        Without Pooling & Audio & 89.74 \\
        \hline
    \end{tabular}
    \label{tab:ablation}
\end{table}
\begin{table*}
\centering

\caption{Comparisons with uni-modal and multi-modal SNNs on CIFAR-10 \& Urbansound  }\label{table:Results-5trials}
\begin{tabular}{c|ccccc}
\hline
Dataset & Methods & Architecture & Timesteps & Modality & Accuracy \\
\hline
\multirow{3}{8em}{CIFAR10} & ~\cite{rathi2020enabling} & ResNet-20 & 250 & uni-modal & $92.22$  \\
 & ~\cite{rathi2020diet} & ResNet-20 & 5 & uni-modal & $92.54$  \\
 &~\cite{zheng2021going} & ResNet-19 & 4 & uni-modal & $92.92$ \\
 &~\cite{deng2022temporal} & ResNet-19 & 4 & uni-modal & $94.44$ \\
 &~\cite{li2021differentiable} & ResNet-18 & 2 & uni-modal & $93.13$ \\ 
 \hline
 \multirow{3}{8em}{CIFAR10-AV} & ~\cite{10293172} & Multi-modal Transformer & 4 & multi-modal & 97.01  \\
 & SMLP & Multi-modal Fusion & 8 & multi-modal & \textbf{98.6}  \\
 \hline
\multirow{3}{8em}{Urbansound8K} & ~\cite{rathi2020diet} & VGG9 & 5 & uni-modal & $95.30$  \\
 & ~\cite{fang2021incorporating} & Cifar10Net & $8$ & uni-modal & $88.60$  \\
& ~\cite{zhou2022spikformer} & Spikformer-4-384 & $4$ & uni-modal & $94.47$  \\
\hline
\multirow{3}{8em}{Urbansound8K-AV} & ~\cite{10293172} & Multi-modal Transformer & $4$ & multi-modal & $96.85$  \\
 & SMLP & Multi-modal transformer & $8$ & multi-modal & \textbf{97.2}  \\

\hline

\end{tabular}
\end{table*}

\subsection{Ablation study}
With a view to evaluate the contributions of different components within our architecture, we conducted a systematic ablation investigation, changing key components, and tracking their effects on accuracy, loss, and computation efficiency throughout training and validation. 
We selected to work on the CIFAR-AV dataset, as it is more complex than the UrbanSound8K-AV dataset.
The parameters evaluated under our ablation study are summarized as follows:

\begin{itemize} 
\item LIF neurons after batch normalization (LIFafterBN): We evaluated the effect of placing Leaky Integrate-and-Fire (LIF) neurons immediately after the batch normalization layer in the visual modality.
\item LIF layer before addition (LIFbeforeADD): We added a LIF layer before the addition operation in both the forward and identity blocks of the visual network to determine its impact on performance. 
\item Removal of the last dropout layer (w/o dropout): In the audio modality, we investigated the effect of removing the final dropout layer.
\item Removal of the 3rd convolutional block (w/o 3rd block): We assessed the impact of excluding the third Conv-BN-LIF-Pool block in the audio network, reducing the network's depth. 
\item Removal of pooling layers (w/o pooling): To evaluate the impact of this modification on feature extraction and classification accuracy, we lastly examined the audio modality's performance in the absence of any pooling layers. 
\end{itemize}

The results of all the conducted experiments are summarized in Table~\ref{tab:ablation}.
The accuracy of $90.40\%$ obtained by placing LIF neurons after batch normalization (LIFafterBN) shows a small decline from the baseline accuracy of $92.74\%$.
This outcome further highlights the importance of our initial proposed approach in maintaining spatio-temporal dynamics.
Furthermore, a LIF layer before addition (LIFbeforeADD) dramatically decreased accuracy to $86.82\%$, indicating that this alteration affects the classification by interfering with the information flow in the residual blocks.
The impressive accuracy of the suggested audio model of $99.60\%$ in the audio modality demonstrates by design efficiency.
An accuracy of $78.16\%$, was obtained by removing the last dropout layer (w/o dropout), most likely as a result of greater overfitting in the absence of regularization.
Inasmuch as accuracy is concerned, this exhibited a significant  drop down to $72.36\%$ when the third convolutional block was removed (w/o 3rd block), highlighting the need for a deeper architecture to extract complex audio information for the CIFAR10-AV.
Last, eliminating pooling layers (w/o pooling) kept accuracy at $89.74\%$, yet it failed to outperform the outcomes of the suggested architecture.

\subsection{Comparative analysis}
Ultimately, we evaluated the performance of our approach against the most advanced unimodal and multi-modal SNNs on the CIFAR-AV and UltraSound8k-AV databases.
The comparison includes both existing SNN-based models and multi-modal methods from the literature, highlighting the improvements in accuracy achieved by our method.
Table~\ref{table:Results-5trials} provides a summary of our comparative study.

We find that several uni-modal techniques, such as those based on ResNet architectures, attain accuracy levels in CIFAR-10 that range from $92.22\%$ to $94.44\%$ \cite{rathi2020enabling, rathi2020diet, zheng2021going, deng2022temporal}.
With an accuracy of $98.6\%$, our multi-modal fusion strategy significantly outperforms the multi-modal transformer-based method in \cite{10293172} achieving $97.01\%$.

In Urbansound8K, uni-modal models such as VGG9 \cite{rathi2020diet} and Spikformer-4-384 \cite{zhou2022spikformer} report accuracies of $95.30\%$ and $94.47\%$, respectively.
Our multi-modal transformer model reaches an accuracy of $97.2\%$, outperforming the multi-modal Transformer-based method in \cite{10293172} that achieves $96.85\%$.

These results demonstrate that our approach not only introduces for the first time high inter-class separability in spiking neural-based approaches but also outperforms the uni-modal SNN models, also leading to superior performance in multi-modal fusion.

\section{Conclusion}
The paper at hand, proposes a novel spiking feature discrimination approach, by introducing an L2 normalization layer with task-specific architectures designed for different modalities, \textit{viz.}, visual and auditory inputs.
Through extensive experimentation, we have capitalized on our motivation regarding the angular concentration of the accumulated feature vectors in an SNN and enhanced performance through the L2 normalization layer after the last hidden layer of the uni-modal architectures.
Subsequently, we have introduced a spiking fusion approach, the SMLP that enhances classification results, while exploiting the feature discrimination from the uni-modal networks.
To that end, the SMPL model effectively leverages the temporal dynamics of the uni-modal networks and further enhances the classification results.
Our proposed approach achieves superior results compared to the existing works in the field.
This approach can drive future studies in spiking feature representation and multi-modal learning processing diverse data modalities.
Lastly, the investigation of feature representation in neuromorphic datasets, the incorporation of this methodology into neuromorphic hardware, and its use in real-time settings constitute some potential fields for future research.
\bibliographystyle{ieeetr}
\bibliography{snn.bib}

\begin{thebibliography}{10}

\bibitem{7723730}
D.~Li, X.~Chen, M.~Becchi, and Z.~Zong, ``Evaluating the energy efficiency of
  deep convolutional neural networks on cpus and gpus,'' in {\em 2016 IEEE
  International Conferences on Big Data and Cloud Computing (BDCloud), Social
  Computing and Networking (SocialCom), Sustainable Computing and
  Communications (SustainCom) (BDCloud-SocialCom-SustainCom)}, pp.~477--484,
  2016.

\bibitem{sze2017efficient}
V.~Sze, Y.-H. Chen, T.-J. Yang, and J.~S. Emer, ``Efficient processing of deep
  neural networks: A tutorial and survey,'' {\em Proceedings of the IEEE},
  vol.~105, no.~12, pp.~2295--2329, 2017.

\bibitem{szegedy2015going}
C.~Szegedy, W.~Liu, Y.~Jia, P.~Sermanet, S.~Reed, D.~Anguelov, D.~Erhan,
  V.~Vanhoucke, and A.~Rabinovich, ``Going deeper with convolutions,'' in {\em
  Proceedings of the IEEE conference on computer vision and pattern
  recognition}, pp.~1--9, 2015.

\bibitem{touvron2021going}
H.~Touvron, M.~Cord, A.~Sablayrolles, G.~Synnaeve, and H.~J{\'e}gou, ``Going
  deeper with image transformers,'' in {\em Proceedings of the IEEE/CVF
  international conference on computer vision}, pp.~32--42, 2021.

\bibitem{he2016deep}
K.~He, X.~Zhang, S.~Ren, and J.~Sun, ``Deep residual learning for image
  recognition,'' in {\em Proceedings of the IEEE conference on computer vision
  and pattern recognition}, pp.~770--778, 2016.

\bibitem{he2015convolutional}
K.~He and J.~Sun, ``Convolutional neural networks at constrained time cost,''
  in {\em Proceedings of the IEEE conference on computer vision and pattern
  recognition}, pp.~5353--5360, 2015.

\bibitem{srivastava2015highway}
R.~K. Srivastava, K.~Greff, and J.~Schmidhuber, ``Highway networks,'' {\em
  arXiv preprint arXiv:1505.00387}, 2015.

\bibitem{shafiq2022deep}
M.~Shafiq and Z.~Gu, ``Deep residual learning for image recognition: A
  survey,'' {\em Applied Sciences}, vol.~12, no.~18, p.~8972, 2022.

\bibitem{zhang2017residual}
K.~Zhang, M.~Sun, T.~X. Han, X.~Yuan, L.~Guo, and T.~Liu, ``Residual networks
  of residual networks: Multilevel residual networks,'' {\em IEEE Transactions
  on Circuits and Systems for Video Technology}, vol.~28, no.~6,
  pp.~1303--1314, 2017.

\bibitem{roy2019towards}
K.~Roy, A.~Jaiswal, and P.~Panda, ``Towards spike-based machine intelligence
  with neuromorphic computing,'' {\em Nature}, vol.~575, no.~7784,
  pp.~607--617, 2019.

\bibitem{hussaini2024spiking}
S.~Hussaini, {\em Spiking Neural Networks for Scalable Visual Place
  Recognition}.
\newblock PhD thesis, Queensland University of Technology, 2024.

\bibitem{gerstner1999spike}
W.~Gerstner, ``The spike response model,'' {\em The Handbook of Biological
  Physics}, vol.~4, pp.~469--516, 1999.

\bibitem{oik2023hybrid}
K.~M. Oikonomou, I.~Kansizoglou, and A.~Gasteratos, ``A hybrid spiking neural
  network reinforcement learning agent for energy-efficient object
  manipulation,'' {\em Machines}, vol.~11, no.~2, p.~162, 2023.

\bibitem{oikonomou2023hybrid}
K.~M. Oikonomou, I.~Kansizoglou, and A.~Gasteratos, ``A hybrid reinforcement
  learning approach with a spiking actor network for efficient robotic arm
  target reaching,'' {\em IEEE Robotics and Automation Letters}, vol.~8, no.~5,
  pp.~3007--3014, 2023.

\bibitem{9340948}
G.~Tang, N.~Kumar, and K.~P. Michmizos, ``Reinforcement co-learning of deep and
  spiking neural networks for energy-efficient mapless navigation with
  neuromorphic hardware,'' in {\em 2020 IEEE/RSJ International Conference on
  Intelligent Robots and Systems (IROS)}, pp.~6090--6097, 2020.

\bibitem{fan2024sfod}
Y.~Fan, W.~Zhang, C.~Liu, M.~Li, and W.~Lu, ``Sfod: Spiking fusion object
  detector,'' in {\em Proceedings of the IEEE/CVF Conference on Computer Vision
  and Pattern Recognition}, pp.~17191--17200, 2024.

\bibitem{5585775}
J.~J. Wade, L.~J. McDaid, J.~A. Santos, and H.~M. Sayers, ``Swat: A spiking
  neural network training algorithm for classification problems,'' {\em IEEE
  Transactions on Neural Networks}, vol.~21, no.~11, pp.~1817--1830, 2010.

\bibitem{ponulak2010supervised}
F.~Ponulak and A.~Kasi{\'n}ski, ``Supervised learning in spiking neural
  networks with resume: sequence learning, classification, and spike
  shifting,'' {\em Neural computation}, vol.~22, no.~2, pp.~467--510, 2010.

\bibitem{9674199}
C.~N. Mavridis and J.~S. Baras, ``Online deterministic annealing for
  classification and clustering,'' {\em IEEE Transactions on Neural Networks
  and Learning Systems}, vol.~34, no.~10, pp.~7125--7134, 2023.

\bibitem{ding2021optimal}
J.~Ding, Z.~Yu, Y.~Tian, and T.~Huang, ``Optimal ann-snn conversion for fast
  and accurate inference in deep spiking neural networks,'' {\em arXiv preprint
  arXiv:2105.11654}, 2021.

\bibitem{10285018}
S.~Liu, Y.~Liang, and Y.~Yi, ``Dnn-snn co-learning for sustainable symbol
  detection in 5g systems on loihi chip,'' {\em IEEE Transactions on
  Sustainable Computing}, vol.~9, no.~2, pp.~170--181, 2024.

\bibitem{yao2023attention}
M.~Yao, G.~Zhao, H.~Zhang, Y.~Hu, L.~Deng, Y.~Tian, B.~Xu, and G.~Li,
  ``Attention spiking neural networks,'' {\em IEEE transactions on pattern
  analysis and machine intelligence}, vol.~45, no.~8, pp.~9393--9410, 2023.

\bibitem{liu2022event}
Q.~Liu, D.~Xing, L.~Feng, H.~Tang, and G.~Pan, ``Event-based multimodal spiking
  neural network with attention mechanism,'' in {\em ICASSP 2022-2022 IEEE
  International Conference on Acoustics, Speech and Signal Processing
  (ICASSP)}, pp.~8922--8926, IEEE, 2022.

\bibitem{oikonomou2023bio}
K.~M. Oikonomou, I.~Kansizoglou, I.~T. Papapetros, and A.~Gasteratos, ``A
  bio-inspired elderly action recognition system for ambient assisted living,''
  in {\em 2023 18th International Workshop on Cellular Nanoscale Networks and
  their Applications (CNNA)}, pp.~1--6, IEEE, 2023.

\bibitem{lee2020enabling}
C.~Lee, S.~S. Sarwar, P.~Panda, G.~Srinivasan, and K.~Roy, ``Enabling
  spike-based backpropagation for training deep neural network architectures,''
  {\em Frontiers in neuroscience}, vol.~14, p.~497482, 2020.

\bibitem{sengupta2019going}
A.~Sengupta, Y.~Ye, R.~Wang, C.~Liu, and K.~Roy, ``Going deeper in spiking
  neural networks: Vgg and residual architectures,'' {\em Frontiers in
  neuroscience}, vol.~13, p.~95, 2019.

\bibitem{shi2024spikingresformer}
X.~Shi, Z.~Hao, and Z.~Yu, ``Spikingresformer: Bridging resnet and vision
  transformer in spiking neural networks,'' in {\em Proceedings of the IEEE/CVF
  Conference on Computer Vision and Pattern Recognition}, pp.~5610--5619, 2024.

\bibitem{shimojo2001sensory}
S.~Shimojo and L.~Shams, ``Sensory modalities are not separate modalities:
  plasticity and interactions,'' {\em Current opinion in neurobiology},
  vol.~11, no.~4, pp.~505--509, 2001.

\bibitem{9746865}
Q.~Liu, D.~Xing, L.~Feng, H.~Tang, and G.~Pan, ``Event-based multimodal spiking
  neural network with attention mechanism,'' in {\em ICASSP 2022 - 2022 IEEE
  International Conference on Acoustics, Speech and Signal Processing
  (ICASSP)}, pp.~8922--8926, 2022.

\bibitem{8482490}
N.~Rathi and K.~Roy, ``Stdp based unsupervised multimodal learning with
  cross-modal processing in spiking neural networks,'' {\em IEEE Transactions
  on Emerging Topics in Computational Intelligence}, vol.~5, no.~1,
  pp.~143--153, 2021.

\bibitem{10293172}
L.~Guo, Z.~Gao, J.~Qu, S.~Zheng, R.~Jiang, Y.~Lu, and H.~Qiao,
  ``Transformer-based spiking neural networks for multimodal audiovisual
  classification,'' {\em IEEE Transactions on Cognitive and Developmental
  Systems}, vol.~16, no.~3, pp.~1077--1086, 2024.

\bibitem{davies2021advancing}
M.~Davies, A.~Wild, G.~Orchard, Y.~Sandamirskaya, G.~A.~F. Guerra, P.~Joshi,
  P.~Plank, and S.~R. Risbud, ``Advancing neuromorphic computing with loihi: A
  survey of results and outlook,'' {\em Proceedings of the IEEE}, vol.~109,
  no.~5, pp.~911--934, 2021.

\bibitem{massa2020efficient}
R.~Massa, A.~Marchisio, M.~Martina, and M.~Shafique, ``An efficient spiking
  neural network for recognizing gestures with a dvs camera on the loihi
  neuromorphic processor,'' in {\em 2020 International Joint Conference on
  Neural Networks (IJCNN)}, pp.~1--9, IEEE, 2020.

\bibitem{iakymchuk2015simplified}
T.~Iakymchuk, A.~Rosado-Mu{\~n}oz, J.~F. Guerrero-Mart{\'\i}nez,
  M.~Bataller-Mompe{\'a}n, and J.~V. Franc{\'e}s-V{\'\i}llora, ``Simplified
  spiking neural network architecture and stdp learning algorithm applied to
  image classification,'' {\em EURASIP Journal on Image and Video Processing},
  vol.~2015, pp.~1--11, 2015.

\bibitem{hao2020biologically}
Y.~Hao, X.~Huang, M.~Dong, and B.~Xu, ``A biologically plausible supervised
  learning method for spiking neural networks using the symmetric stdp rule,''
  {\em Neural Networks}, vol.~121, pp.~387--395, 2020.

\bibitem{kheradpisheh2018stdp}
S.~R. Kheradpisheh, M.~Ganjtabesh, S.~J. Thorpe, and T.~Masquelier,
  ``Stdp-based spiking deep convolutional neural networks for object
  recognition,'' {\em Neural Networks}, vol.~99, pp.~56--67, 2018.

\bibitem{bu2023optimal}
T.~Bu, W.~Fang, J.~Ding, P.~Dai, Z.~Yu, and T.~Huang, ``Optimal ann-snn
  conversion for high-accuracy and ultra-low-latency spiking neural networks,''
  {\em arXiv preprint arXiv:2303.04347}, 2023.

\bibitem{wang2024universal}
Y.~Wang, H.~Liu, M.~Zhang, X.~Luo, and H.~Qu, ``A universal ann-to-snn
  framework for achieving high accuracy and low latency deep spiking neural
  networks,'' {\em Neural Networks}, vol.~174, p.~106244, 2024.

\bibitem{wang2023new}
B.~Wang, J.~Cao, J.~Chen, S.~Feng, and Y.~Wang, ``A new ann-snn conversion
  method with high accuracy, low latency and good robustness.,'' in {\em
  IJCAI}, pp.~3067--3075, 2023.

\bibitem{deng2022temporal}
S.~Deng, Y.~Li, S.~Zhang, and S.~Gu, ``Temporal efficient training of spiking
  neural network via gradient re-weighting,'' {\em arXiv preprint
  arXiv:2202.11946}, 2022.

\bibitem{guo2023membrane}
Y.~Guo, Y.~Zhang, Y.~Chen, W.~Peng, X.~Liu, L.~Zhang, X.~Huang, and Z.~Ma,
  ``Membrane potential batch normalization for spiking neural networks,'' in
  {\em Proceedings of the IEEE/CVF International Conference on Computer
  Vision}, pp.~19420--19430, 2023.

\bibitem{meng2023towards}
Q.~Meng, M.~Xiao, S.~Yan, Y.~Wang, Z.~Lin, and Z.-Q. Luo, ``Towards memory-and
  time-efficient backpropagation for training spiking neural networks,'' in
  {\em Proceedings of the IEEE/CVF International Conference on Computer
  Vision}, pp.~6166--6176, 2023.

\bibitem{wei2023temporal}
W.~Wei, M.~Zhang, H.~Qu, A.~Belatreche, J.~Zhang, and H.~Chen, ``Temporal-coded
  spiking neural networks with dynamic firing threshold: Learning with
  event-driven backpropagation,'' in {\em Proceedings of the IEEE/CVF
  International Conference on Computer Vision}, pp.~10552--10562, 2023.

\bibitem{zheng2021going}
H.~Zheng, Y.~Wu, L.~Deng, Y.~Hu, and G.~Li, ``Going deeper with
  directly-trained larger spiking neural networks,'' in {\em Proceedings of the
  AAAI conference on artificial intelligence}, vol.~35, pp.~11062--11070, 2021.

\bibitem{fang2021deep}
W.~Fang, Z.~Yu, Y.~Chen, T.~Huang, T.~Masquelier, and Y.~Tian, ``Deep residual
  learning in spiking neural networks,'' {\em Advances in Neural Information
  Processing Systems}, vol.~34, pp.~21056--21069, 2021.

\bibitem{zhou2023spikingformer}
C.~Zhou, L.~Yu, Z.~Zhou, Z.~Ma, H.~Zhang, H.~Zhou, and Y.~Tian,
  ``Spikingformer: Spike-driven residual learning for transformer-based spiking
  neural network,'' {\em arXiv preprint arXiv:2304.11954}, 2023.

\bibitem{hu2021spiking}
Y.~Hu, H.~Tang, and G.~Pan, ``Spiking deep residual networks,'' {\em IEEE
  Transactions on Neural Networks and Learning Systems}, vol.~34, no.~8,
  pp.~5200--5205, 2021.

\bibitem{natarajan2012multimodal}
P.~Natarajan, S.~Wu, S.~Vitaladevuni, X.~Zhuang, S.~Tsakalidis, U.~Park,
  R.~Prasad, and P.~Natarajan, ``Multimodal feature fusion for robust event
  detection in web videos,'' in {\em 2012 IEEE Conference on Computer Vision
  and Pattern Recognition}, pp.~1298--1305, IEEE, 2012.

\bibitem{ebrahimi2015recurrent}
S.~Ebrahimi~Kahou, V.~Michalski, K.~Konda, R.~Memisevic, and C.~Pal,
  ``Recurrent neural networks for emotion recognition in video,'' in {\em
  Proceedings of the 2015 ACM on international conference on multimodal
  interaction}, pp.~467--474, 2015.

\bibitem{metallinou2012context}
A.~Metallinou, M.~Wollmer, A.~Katsamanis, F.~Eyben, B.~Schuller, and
  S.~Narayanan, ``Context-sensitive learning for enhanced audiovisual emotion
  classification,'' {\em IEEE Transactions on Affective Computing}, vol.~3,
  no.~2, pp.~184--198, 2012.

\bibitem{eyben2011audiovisual}
F.~Eyben, S.~Petridis, B.~Schuller, G.~Tzimiropoulos, S.~Zafeiriou, and
  M.~Pantic, ``Audiovisual classification of vocal outbursts in human
  conversation using long-short-term memory networks,'' in {\em 2011 IEEE
  International Conference on Acoustics, Speech and Signal Processing
  (ICASSP)}, pp.~5844--5847, IEEE, 2011.

\bibitem{8937495}
I.~Kansizoglou, L.~Bampis, and A.~Gasteratos, ``An active learning paradigm for
  online audio-visual emotion recognition,'' {\em IEEE Transactions on
  Affective Computing}, vol.~13, no.~2, pp.~756--768, 2022.

\bibitem{feng2017audio}
W.~Feng, N.~Guan, Y.~Li, X.~Zhang, and Z.~Luo, ``Audio visual speech
  recognition with multimodal recurrent neural networks,'' in {\em 2017
  International Joint Conference on neural networks (IJCNN)}, pp.~681--688,
  IEEE, 2017.

\bibitem{makino2019recurrent}
T.~Makino, H.~Liao, Y.~Assael, B.~Shillingford, B.~Garcia, O.~Braga, and
  O.~Siohan, ``Recurrent neural network transducer for audio-visual speech
  recognition,'' in {\em 2019 IEEE automatic speech recognition and
  understanding workshop (ASRU)}, pp.~905--912, IEEE, 2019.

\bibitem{rathi2018stdp}
N.~Rathi and K.~Roy, ``Stdp based unsupervised multimodal learning with
  cross-modal processing in spiking neural networks,'' {\em IEEE Transactions
  on Emerging Topics in Computational Intelligence}, vol.~5, no.~1,
  pp.~143--153, 2018.

\bibitem{sun2014deep}
Y.~Sun, Y.~Chen, X.~Wang, and X.~Tang, ``Deep learning face representation by
  joint identification-verification,'' {\em Advances in neural information
  processing systems}, vol.~27, 2014.

\bibitem{wen2016discriminative}
Y.~Wen, K.~Zhang, Z.~Li, and Y.~Qiao, ``A discriminative feature learning
  approach for deep face recognition,'' in {\em Computer vision--ECCV 2016:
  14th European conference, amsterdam, the netherlands, October 11--14, 2016,
  proceedings, part VII 14}, pp.~499--515, Springer, 2016.

\bibitem{adeli2018semi}
E.~Adeli, K.-H. Thung, L.~An, G.~Wu, F.~Shi, T.~Wang, and D.~Shen,
  ``Semi-supervised discriminative classification robust to sample-outliers and
  feature-noises,'' {\em IEEE transactions on pattern analysis and machine
  intelligence}, vol.~41, no.~2, pp.~515--522, 2018.

\bibitem{liu2016large}
W.~Liu, Y.~Wen, Z.~Yu, and M.~Yang, ``Large-margin softmax loss for
  convolutional neural networks,'' {\em arXiv preprint arXiv:1612.02295}, 2016.

\bibitem{liu2017deep}
W.~Liu, Y.-M. Zhang, X.~Li, Z.~Yu, B.~Dai, T.~Zhao, and L.~Song, ``Deep
  hyperspherical learning,'' {\em Advances in neural information processing
  systems}, vol.~30, 2017.

\bibitem{deng2019arcface}
J.~Deng, J.~Guo, N.~Xue, and S.~Zafeiriou, ``Arcface: Additive angular margin
  loss for deep face recognition,'' in {\em Proceedings of the IEEE/CVF
  conference on computer vision and pattern recognition}, pp.~4690--4699, 2019.

\bibitem{liu2017sphereface}
W.~Liu, Y.~Wen, Z.~Yu, M.~Li, B.~Raj, and L.~Song, ``Sphereface: Deep
  hypersphere embedding for face recognition,'' in {\em Proceedings of the IEEE
  conference on computer vision and pattern recognition}, pp.~212--220, 2017.

\bibitem{9477034}
I.~Kansizoglou, L.~Bampis, and A.~Gasteratos, ``Deep feature space: A
  geometrical perspective,'' {\em IEEE Transactions on Pattern Analysis and
  Machine Intelligence}, vol.~44, no.~10, pp.~6823--6838, 2022.

\bibitem{paszke2017automatic}
A.~Paszke, S.~Gross, S.~Chintala, G.~Chanan, E.~Yang, Z.~DeVito, Z.~Lin,
  A.~Desmaison, L.~Antiga, and A.~Lerer, ``Automatic differentiation in
  pytorch,'' in {\em NIPS-W}, 2017.

\bibitem{kingma2014adam}
D.~P. Kingma, ``Adam: A method for stochastic optimization,'' {\em arXiv
  preprint arXiv:1412.6980}, 2014.

\bibitem{fang2021incorporating}
W.~Fang, Z.~Yu, Y.~Chen, T.~Masquelier, T.~Huang, and Y.~Tian, ``Incorporating
  learnable membrane time constant to enhance learning of spiking neural
  networks,'' in {\em Proceedings of the IEEE/CVF international conference on
  computer vision}, pp.~2661--2671, 2021.

\bibitem{rathi2020enabling}
N.~Rathi, G.~Srinivasan, P.~Panda, and K.~Roy, ``Enabling deep spiking neural
  networks with hybrid conversion and spike timing dependent backpropagation,''
  {\em arXiv preprint arXiv:2005.01807}, 2020.

\bibitem{rathi2020diet}
N.~Rathi and K.~Roy, ``Diet-snn: Direct input encoding with leakage and
  threshold optimization in deep spiking neural networks,'' {\em arXiv preprint
  arXiv:2008.03658}, 2020.

\bibitem{li2021differentiable}
Y.~Li, Y.~Guo, S.~Zhang, S.~Deng, Y.~Hai, and S.~Gu, ``Differentiable spike:
  Rethinking gradient-descent for training spiking neural networks,'' {\em
  Advances in Neural Information Processing Systems}, vol.~34,
  pp.~23426--23439, 2021.

\bibitem{zhou2022spikformer}
Z.~Zhou, Y.~Zhu, C.~He, Y.~Wang, S.~Yan, Y.~Tian, and L.~Yuan, ``Spikformer:
  When spiking neural network meets transformer,'' {\em arXiv preprint
  arXiv:2209.15425}, 2022.

\end{thebibliography}





 



\begin{IEEEbiography}[{\includegraphics[width=1in,height=1.25in,clip,keepaspectratio]{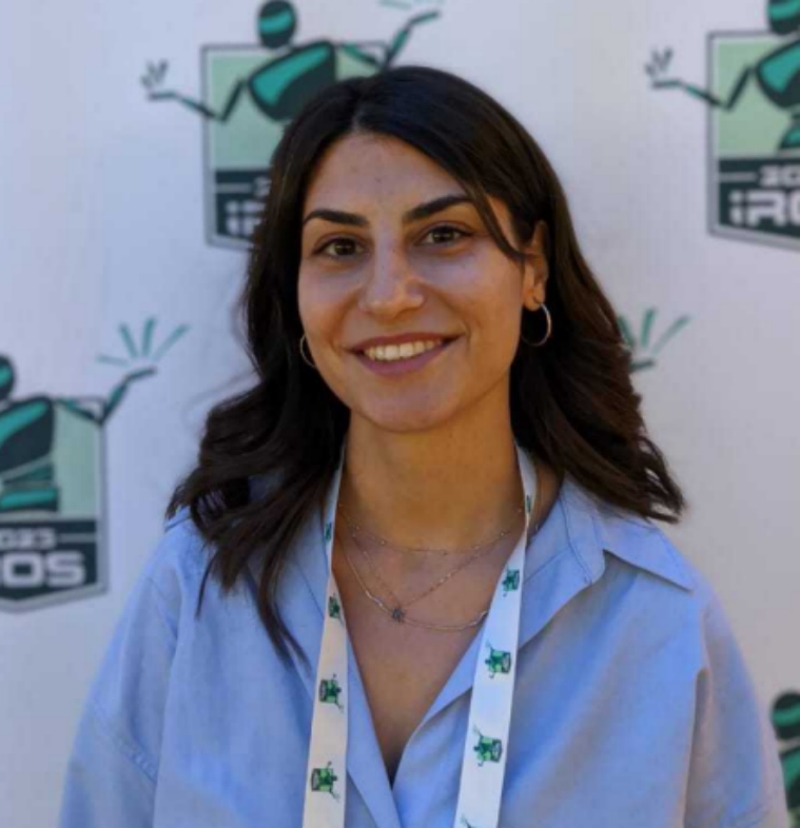}}]{Oikonomou Katerina Maria} is a Ph.D. Candidate at the Democritus University of Thrace, Xanthi, Greece, working with Prof. Antonios Gasteratos. She received her Masters’s Degree in Robotics and Automation Systems from the National Technical University of Athens, Greece, having her Master’s thesis achieved at the University of Bremen, Germany. She also holds a Bachelor's Degree in Physics from the University of Patras, Greece, having followed the informatics, electronics, and signal processing specialization. Her research interests include robotics, human-robot interaction, and spiking neural networks. More details about her are available at: 
https://robotics.pme.duth.gr/koikonomou/
\end{IEEEbiography}

\begin{IEEEbiography}[{\includegraphics[width=1in,height=1.25in,clip,keepaspectratio]{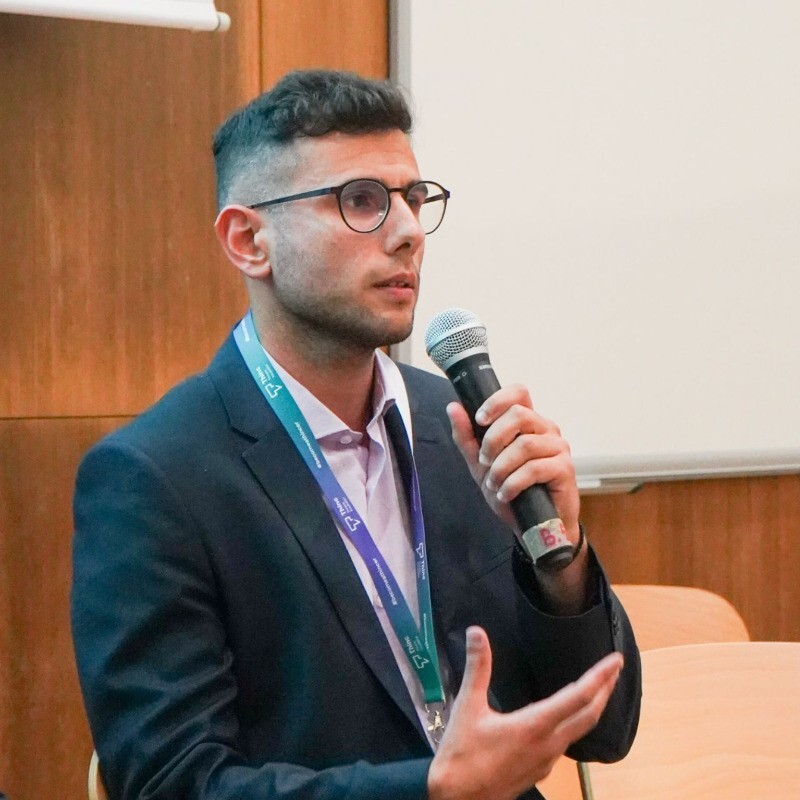}}]{Ioannis Kansizoglou} received the Diploma degree in electrical and computer engineering from the Aristotle University of Thessaloniki, Thessaloniki, Greece, in 2017, and the Ph.D. degree in deep representation learning and computer vision from the Laboratory of Robotics and Automation, Department of Production and Management Engineering, Democritus University of Thrace, Xanthi, Greece, in 2021.
He is a Post-Doctoral Researcher with the Laboratory of Robotics and Automation, Department of Production and Management Engineering, Democritus University of Thrace. His research interests include deep representation learning, emotion analysis, and human-robot interaction. His work is supported by several research projects funded by the European Commission and the Greek Government.
\end{IEEEbiography}
\begin{IEEEbiography}[{\includegraphics[width=1in,height=1.25in, clip, keepaspectratio]{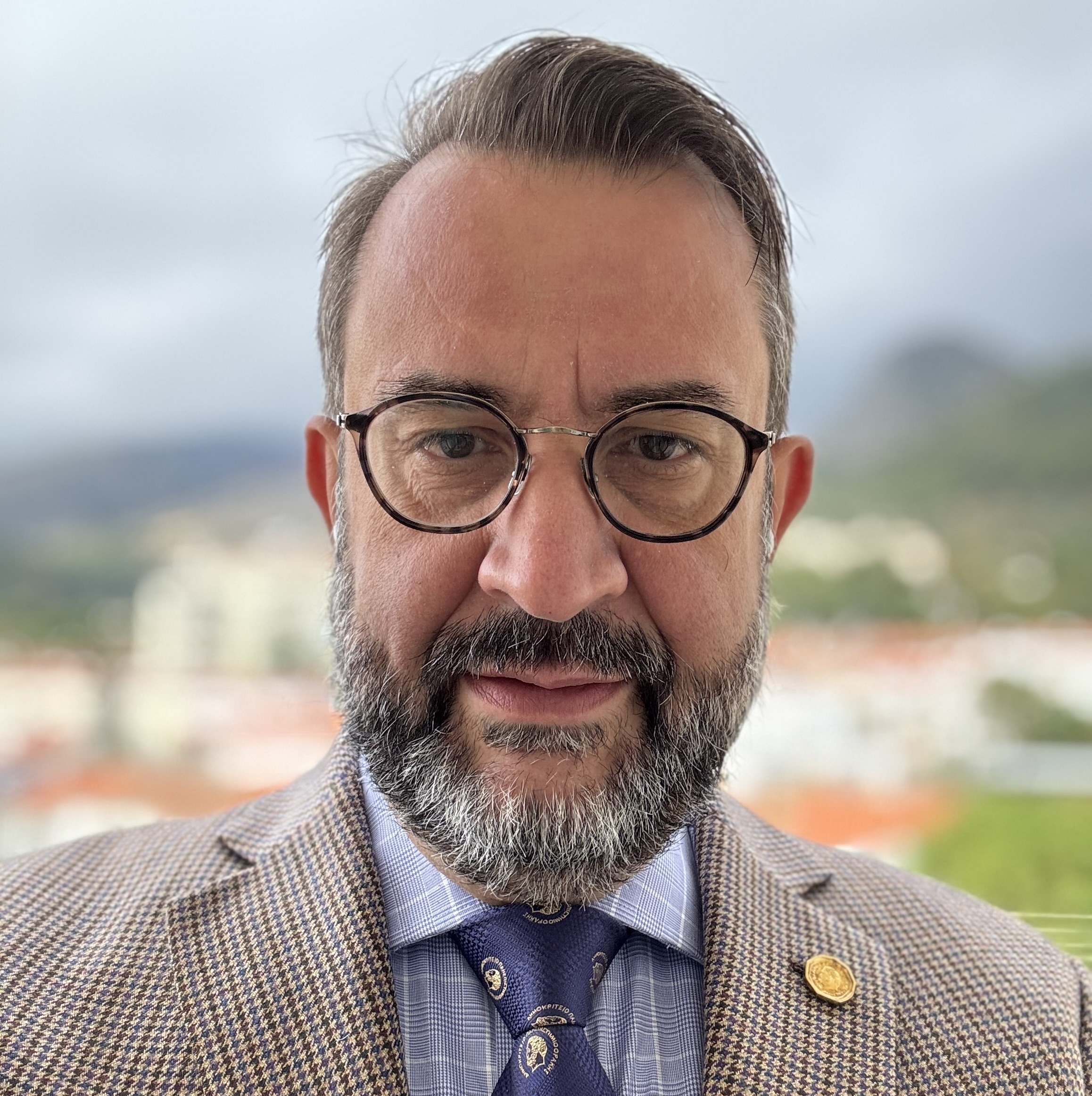}}]{Antonios Gasteratos} (Senior Member, IEEE) received the M.Eng. and Ph.D. degrees from the Department of Electrical and Computer Engineering, Democritus University of Thrace (DUTh), Xanthi, Greece, in 1994 and 1998, respectively.
From 1999 to 2000, he was a Visiting Researcher with the Laboratory of Integrated Advanced Robotics (LIRALab), DIST, University of Genoa, Genoa, Italy. He is currently a Professor and the Head of the Production and Management Engineering Department at DUTH.
He is also the Director of the Laboratory of Robotics and Automation at DUTH and teaches courses in robotics, automatic control systems, electronics, mechatronics, and computer vision. He has authored over 220 papers in books, journals, and conferences.
His research interests include mechatronics and robot vision. Dr. Gasteratos is a Fellow member of IET.
He has served as a reviewer for numerous scientific journals and international conferences.
He is a Subject Editor of Electronics Letters and an Associate Editor of the International Journal of Optomecatronics.
He has organized/co-organized several international conferences. More details about him are available at:
http://robotics.pme.duth.gr/antonis.
\end{IEEEbiography}

\vfill

\end{document}